\documentclass{article}

\usepackage[final]{neurips_2025}

\usepackage[utf8]{inputenc} 
\usepackage[T1]{fontenc}    
\usepackage{hyperref}       
\usepackage{url}            
\usepackage{booktabs}       
\usepackage{amsfonts}       
\usepackage{nicefrac}       
\usepackage{microtype}      
\usepackage{xcolor}         
\usepackage{algorithm}
\usepackage{algpseudocode}
\usepackage{graphicx}
\usepackage{enumitem}
\usepackage{tikz}
\usepackage{subcaption}

\usetikzlibrary{arrows.meta}

\usepackage{amsmath,amsfonts,bm}

\def\eqref#1{equation~\ref{#1}}

\def\1{\bm{1}}

\DeclareMathAlphabet{\mathsfit}{\encodingdefault}{\sfdefault}{m}{sl}
\SetMathAlphabet{\mathsfit}{bold}{\encodingdefault}{\sfdefault}{bx}{n}

\def\sA{{\mathbb{A}}}
\def\sB{{\mathbb{B}}}

\def\sD{{\mathbb{D}}}

\def\sI{{\mathbb{I}}}

\def\sP{{\mathbb{P}}}

\newcommand{\RankRefine}{\texttt{RankRefine}}

\newtheorem{theorem}{Theorem}[section]
\newtheorem{corollary}{Corollary}[theorem]
\newtheorem{lemma}[theorem]{Lemma}

\title{Post Hoc Regression Refinement via Pairwise Rankings}

\author{%
  Kevin Tirta Wijaya \\
  MPI-INF \\
  \texttt{kevintirta.w@gmail.com} \\
  \And
  Michael Sun \\
  MIT \\
  \texttt{msun415@csail.mit.edu}\\
  \And
  Minghao Guo \\
  MIT \\
  \texttt{guomh2014@gmail.com}\\
  \AND
  Hans-Peter Seidel \\
  MPI-INF \\
  \texttt{hpseidel@mpi-inf.mpg.de} \\
  \And
  Wojciech Matusik\\
  MIT \\
  \texttt{wojciech@csail.mit.edu} \\
  \And
  Vahid Babaei \\
  MPI-INF \\
  \texttt{vbabaei@mpi-inf.mpg.de}
}

\begin{document}

\maketitle

\begin{abstract}
Accurate prediction of continuous properties is essential to many scientific and engineering tasks. 
Although deep-learning regressors excel with abundant labels, their accuracy deteriorates in data-scarce regimes. 
We introduce \RankRefine, a model-agnostic, plug-and-play post hoc method that refines regression with expert knowledge coming from pairwise rankings. 
Given a query item and a small reference set with known properties, \RankRefine\ combines the base regressor’s output with a rank-based estimate via inverse-variance weighting, requiring no retraining. 
In molecular property prediction task, \RankRefine\ achieves up to 10\% relative reduction in mean absolute error using only 20 pairwise comparisons obtained through a general-purpose large language model (LLM) with no finetuning.
As rankings provided by human experts or general-purpose LLMs are sufficient for improving regression across diverse domains, \RankRefine\ offers practicality and broad applicability, especially in low-data settings.
\end{abstract}
\section{Introduction}
Accurate prediction of continuous properties is crucial across scientific and engineering disciplines.
In molecular‐property prediction (MPP), for instance, reliable estimates of physical or chemical attributes can accelerate drug discovery, materials design, and catalyst development.
Recent advances in deep learning have enhanced regression models by leveraging large datasets to uncover complex, nonlinear relationships between structured inputs (e.g., molecular graphs, crystal structures) and their associated properties.
Yet, unlike computer vision or natural-language processing, where large labeled corpora can be mined or crowd-sourced, many specialized fields face a fundamental bottleneck: acquiring ground-truth labels requires expert-led experiments that are both costly and slow.
Consequently, real-world tasks often operate in data-scarce regimes where even 50 labeled samples may be considered plentiful \citep{sun2024representing}.

When large-scale labels are unavailable, expert knowledge becomes a valuable but under-utilized asset.
One accessible form of such knowledge is relative comparison: pairwise judgments about which of two samples exhibits a higher (or lower) property value.
Pairwise rankings convey substantial information, are often easier for human experts to provide, and can even be generated by general-purpose large language models (LLMs), which have shown surprising ranking skills in chemistry and related domains \citep{sun2025fmg, guo2023can_llmforchemistry}.

We introduce \RankRefine, a post hoc refinement framework that improves regression predictions by incorporating pairwise rankings.
Given a query sample and a small set of labeled references, which can conveniently be drawn from training data, a ranker infers the relative ordering between the query and each reference; these comparisons define a likelihood whose minimization yields a rank-based property estimate.
\RankRefine\ then fuses this estimate with the base regressor’s prediction via inverse-variance weighting, all without retraining.

Because \RankRefine\ is largely model-agnostic and requires no architectural changes, it suits low-data, few-shot, and meta-learning scenarios.
Leveraging ranking signals from publicly hosted LLMs, it provides substantial improvements in regression accuracy with minimal labeled data and negligible client-side computation.
Experiments on synthetic and real-world benchmarks, including multiple MPP tasks, demonstrate that \RankRefine\ consistently boosts predictive performance.
Further analysis links its effectiveness to ranking quality, underscoring the promise of LLMs and other expert-informed rankers as practical complements to data-scarce regression models.
The source code is available at \hyperlink{https://github.com/ktirta/regression-refinement}{https://github.com/ktirta/regref}.
\section{Background}
\subsection{Combining Regression and Ranking}
\citet{huang2024rankup_regression_ranking} reformulate regression as joint regression–pairwise-ranking objectives and show that the combination improves predictive accuracy.
\citet{tynes2021pairwise_padre,fralish2023deepdelta} instead train a regressor to predict pairwise differences in molecular properties; at inference, a query molecule is compared with reference molecules of known labels, and the mean of the predicted differences yields the final estimate.
The post hoc method proposed by \citet{gonccalves2023regression_regressionbyreranking} refines regression outputs by computing a weighted average of the original prediction and those of its top-k nearest neighbors.

A closely related approach is the consolidation method for document relevance proposed by \citet{yan2024consolidating_projection}.
Given a pointwise rating and a set of pairwise comparisons, their post-processor makes minimal adjustments to preserve the ranker-derived inequalities.
For a one-dimensional rating $\hat{y}$ this projection simplifies to
\[
\hat{y}^{*}= \max\!\bigl(\min(\hat{y},U),L\bigr),
\]
where $L$ and $U$ are the lower and upper bounds of the feasible interval.

Preference learning with Gaussian processes \citep{chu2005preference_preference_learning,houlsby2012collaborative_preference_learning,chau2022learning_preference_learning} utilize pairwise data but differ fundamentally from our setting.
Preference learning assumes a latent score $f(x)$ whose absolute scale is irrelevant; the aim is to preserve ordering and learn-to-rank, not to predict calibrated values.
In contrast, we target real-world quantities, e.g., solubility or toxicity, and treat pairwise comparisons as auxiliary signals that refine supervised regression into meaningful, calibrated estimates.
Similarly, many learning-to-rank approaches \citep{burges2005learning_rankingRegression, yildiz2020fast_rankingRegression, tom2024ranking_rankingRegression} may seem related to our work, but they actually address a different problem, i.e., ranking rather than regression.

\subsection{LLM as a Pairwise Ranker}
Large language models (LLMs) have recently been explored in molecular-property prediction and generation \citep{sun2025fmg}.
Several efforts fine-tune GPT-3- or LLaMA-based models with chemistry-specific prompts \citep{jablonka2024leveraging_llm_finetuned_molecule,xie2024fine_llm_finetuned_molecule,jacobs2024regression_llm_finetuned_molecule}, while MolecularGPT \citep{liu2024moleculargpt} is instruction-tuned on 1\,000 property-prediction tasks and achieves few-shot performance competitive with graph neural networks.

Fine-tuning, however, is resource-intensive and incurs substantial computational overhead.
We therefore ask: can general-purpose, publicly hosted LLMs be leveraged directly, without domain-specific adaptation?
We hypothesize that such models can supply useful signals when deployed as pairwise rankers.

Human cognition offers an analogue.
People often judge quantities either by estimating absolute magnitudes (regression) or by deciding which of two items is greater (ranking).
A long-standing hypothesis holds that comparative judgments are easier and less biased \citep{thurstone1927law_human_better_ranker,miller1956magical_human_better_ranker,stewart2005absolute_human_better_ranker}.
Pairwise tasks impose lower cognitive load \citep{routh2023rating_ranking_lower_cognitive_load} and avoid scale-interpretation bias \citep{hoeijmakers2024subjective_ranker_less_subjective}.
Because LLMs are pretrained on corpora rich in relative statements and later aligned with human feedback \citep{zhang2023instruction_sft,bai2022training_rlhf,rafailov2023direct_dpo}, they may inherit a similar advantage.
As a real example, \citet{guo2023can_llmforchemistry} show that off-the-shelf LLMs rival domain-specific models on chemistry classification and ranking benchmarks, underscoring the promise of pairwise prompting in specialized tasks.

\section{Method}
\begin{figure}[t]
    \centering
    \includegraphics[width=0.99\linewidth]{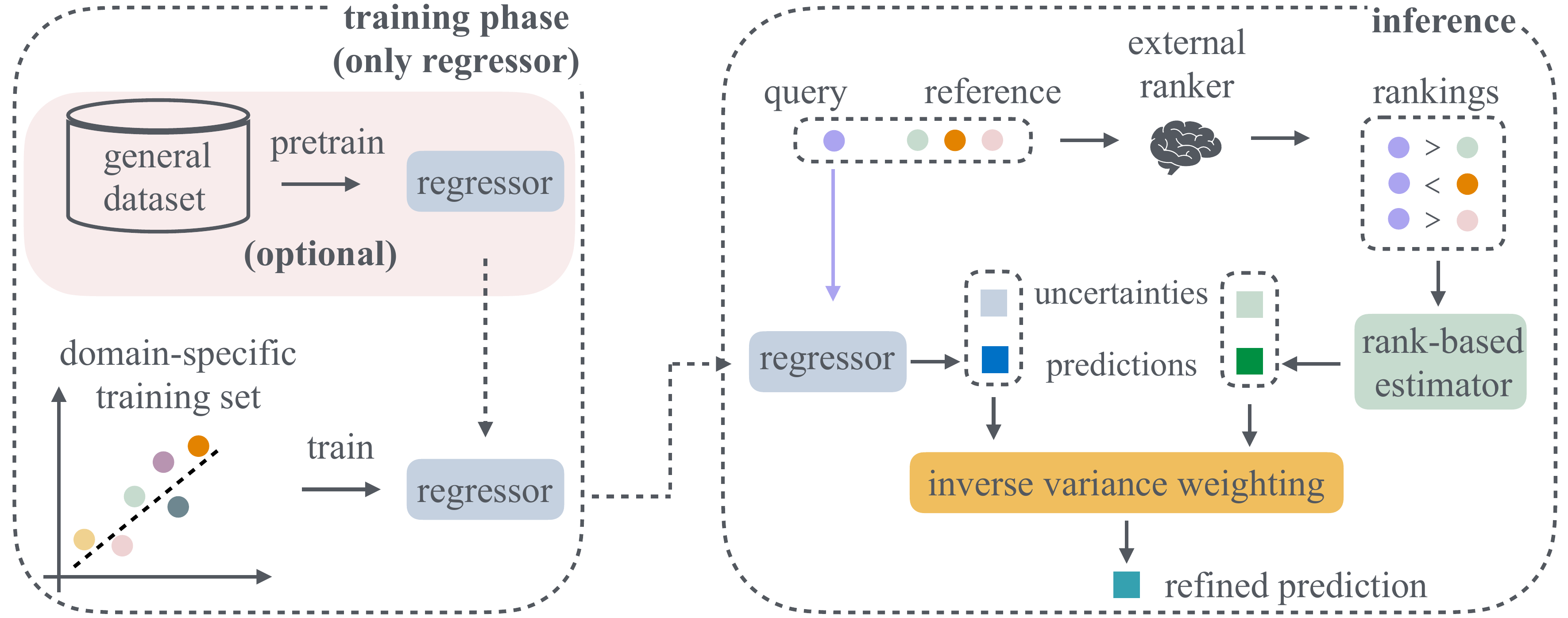}
    \caption{Overview of the \RankRefine\ framework. During training, a regressor is trained using labeled data. At inference, a query sample is paired with reference samples with known properties and compared by an external ranker to produce pairwise rankings. These rankings are used to estimate the query’s property value via a rank-based estimator. The final prediction is obtained by fusing the regressor’s output and the rank-based estimate using inverse variance weighting.}
    \label{fig:rankrefine}
\end{figure}
Our goal is to reduce the test-time error of an existing regressor by fusing its prediction with pairwise-ranking information in a plug-and-play fashion.
We present \RankRefine, a framework that augments a base regressor with rankings from an external source (e.g., an LLM or a domain expert).
\RankRefine\ is largely model-agnostic and applies to any regressor that returns both a point estimate and an uncertainty measure, for example Gaussian processes or random forests.

We assume access to a reference set $\sD=\{(x_i,y_i)\}_{i=1}^{k}$ with known property values.
This reference set can be drawn from the training set.
At inference, the regressor predicts the property of a query sample $x_0$, giving $\hat{y}_0^{\text{reg}}$, while the ranker supplies pairwise comparisons between $x_0$ and each $x_i\in\sD$.

\subsection{\RankRefine}
Let $\hat{y}_0^{\text{reg}}\sim\mathcal{N}(y_0,\sigma_{\text{reg}}^{2})$ be the regressor’s unbiased Gaussian estimate of the true property $y_0$.
A second, rank-based estimate $\hat{y}_0^{\text{rank}}$ is derived from the comparisons.
Under these assumptions, the optimal fusion of the two estimates reduces to the inverse-variance weighting \citep{cochran1953sampling_best_combination_ivm}, formalized below.

\begin{theorem}[RankRefine Fusion Theorem]
\label{thm:rankrefine}
If $\hat{y}_0^{\text{reg}}$ and $\hat{y}_0^{\text{rank}}$ are independent, unbiased Gaussian estimators of $y_0$ with variances $\sigma_{\text{reg}}^{2}$ and $\sigma_{\text{rank}}^{2}$, the minimum-variance unbiased estimator is
\begin{equation}
\hat{y}_0^{*}= \sigma_{\text{post}}^{2}\!\left(
\frac{\hat{y}_0^{\text{reg}}}{\sigma_{\text{reg}}^{2}}
+\frac{\hat{y}_0^{\text{rank}}}{\sigma_{\text{rank}}^{2}}
\right),\qquad
\sigma_{\text{post}}^{2}= \left(
\frac{1}{\sigma_{\text{reg}}^{2}}
+\frac{1}{\sigma_{\text{rank}}^{2}}
\right)^{-1}.
\label{eq:ivw}
\end{equation}
\end{theorem}
See supplementary materials for the proof.

\paragraph{Rank-based estimate.}
Pairwise probabilities follow the Bradley–Terry model \citep{bradley1952rank_bradley_terry},
$P(x_i\succ x_j)=s(y_i-y_j)$ with sigmoid function $s(z)=\bigl(1+e^{-z}\bigr)^{-1}$.
Indexing the query as $i=0$ and references as $i>0$, we obtain $\hat{y}_0^{\text{rank}}$ by minimizing the negative log-likelihood
\begin{equation}
\hat{y}_0^{*\text{rank}}
=\arg\min_{\hat{y}_0^{\text{rank}}}\!
\Bigl[-\!\!\sum_{x_i\in\sA}\!\log s(\hat{y}_0^{\text{rank}}-y_i)
-\!\!\sum_{x_j\in\sB}\!\log\!\bigl(1-s(\hat{y}_0^{\text{rank}}-y_j)\bigr)\Bigr],
\label{eq:nll}
\end{equation}
where $\sA$ contains references ranked below $x_0$ and $\sB$ those ranked above.

\begin{lemma}[Variance of the rank-based estimate]
\label{lem:rank_variance}
Let $\hat{y}_0^{*\text{rank}}$ minimize \eqref{eq:nll}.  
Its variance is approximated by the inverse observed Fisher information \citep{ly2017tutorial},
\begin{equation}
\sigma_{\text{rank}}^{2}\approx
\Bigl[\sum_{y_i\in\sA\cup\sB}
s(\Delta_i)\bigl(1-s(\Delta_i)\bigr)\Bigr]^{-1},
\quad\text{with } \Delta_i=\hat{y}_0^{*\text{rank}}-y_i.
\label{eq:sigma_rank_estimate}
\end{equation}
\end{lemma}

Applying Theorem~\ref{thm:rankrefine} with $\sigma_{\text{rank}}^{2}$ from Lemma~\ref{lem:rank_variance} yields $\hat{y}_0^{*}$.

\subsection{Analysis of \RankRefine}

We measure performance via the mean absolute error (MAE).  
For a folded Gaussian derived from a zero-mean Gaussian, $\text{MAE}=\sqrt{2/\pi}\,\sigma$, so
\[
\text{MAE}_{\text{post}}<\text{MAE}_{\text{reg}}
\;\iff\;
\sigma_{\text{post}}^{2}<\sigma_{\text{reg}}^{2}.
\]
\begin{corollary}
\label{cor:any_informative_ranker}
Any informative ranker with finite variance ($\sigma_{\text{rank}}^{2}<\infty$) lowers the expected MAE after fusion.
\end{corollary}

More generally, letting $\alpha\in[0,1]$ be the desired ratio between post-refinement and the original MAEs,
\begin{equation}   \text{MAE}_{\text{post}}\le\alpha\,\text{MAE}_{\text{reg}}
    \;\Longleftarrow\;
    \sigma_{\text{rank}}^{2}\le
    \frac{\alpha^{2}\sigma_{\text{reg}}^{2}}{1-\alpha^{2}}.
    \label{eq:beta_bound}
\end{equation}

\paragraph{Regularization.}
If the ranker is biased yet over-confident ($\sigma_{\text{rank}}^{2}\ll\sigma_{\text{reg}}^{2}$), we temper its variance via
\[
\sigma_{\text{rank}}^{2}\leftarrow
\max\!\bigl(\sigma_{\text{rank}}^{2},\,c\,\sigma_{\text{reg}}^{2}\bigr),
\]
with user-chosen constant $c>0$.

\section{Experimental Results}
We evaluate \RankRefine\ on synthetic and real-world tasks.  
After outlining datasets, metrics, and implementation details, we report results in synthetic settings, where ranking oracle is available, and practical settings across multiple domains.

\subsection{Experimental Setup} 
We use nine molecular datasets from the TDC ADME benchmark \citep{huang2021therapeutics_tdcDataset}:  
Caco-2 \citep{wang2016adme_caco2}, Clearance Microsome and Clearance Hepatocyte \citep{di2012mechanistic_clearance}, log Half-Life \citep{obach2008trend_half_life}, FreeSolv \citep{mobley2014freesolv}, Lipophilicity \citep{wu2018moleculenet_lipo}, PPBR, Solubility \citep{sorkun2019aqsoldb_solubility}, and VDss \citep{lombardo2016silico_vdss}.  
We additionally test three tabular regressions: crop-yield prediction from sensor data \citep{misc_smart_farming}, student-performance prediction \citep{misc_student_performance_320}, international-education cost estimation \citep{misc_education_cost}.
In the human-as-ranker experiment, we use UTKFace \citep{zhang2017age_utkface} for age estimation.

To emulate low-data regimes, we sample 50 training points from above datasets uniformly at random and merge the remainder with the original test split, repeating this re-split over five random seeds.

The primary metric is the mean absolute error (MAE).  
We report the \textit{normalized error}  
\[
\beta=\frac{\text{MAE}_{\text{post}}}{\text{MAE}_{\text{reg}}},
\]
where $\beta<1$ indicates improvement.  

Suppose that a pairwise ranker $R$ evaluates a pair of inputs $(x_i, x_j)$.
The ranker returns $R(x_i,x_j) = 0$ if it predicts $x_i \succ x_j$, and $R(x_i, x_j) = 1$ otherwise.
The pairwise ranker quality can be measured via pairwise ranking accuracy (PRA), defined as,
\[
\text{PRA} = \frac{1}{|\sP|} \sum_{(i,j) \in \sP}\sI\bigl(R(x_i, x_j) = \sI(y_i < y_j)\bigr),
\]
where $\sP = \{(i, j) | i\neq j\}$ is the set of all comparable pairs with  ground-truth labels $y_i, y_j$, and $\sI$ is the indicator function which return 1 if the condition is true, and 0 otherwise.
Unless stated otherwise, the base model is a random-forest regressor from scikit-learn \citep{pedregosa2011scikit} with default hyper-parameters, executed on a single CPU.

\subsection{Verifying Theoretical MAE Reduction Under Ideal Conditions}
\begin{figure}[t]
    \centering
\includegraphics[width=0.5\linewidth]{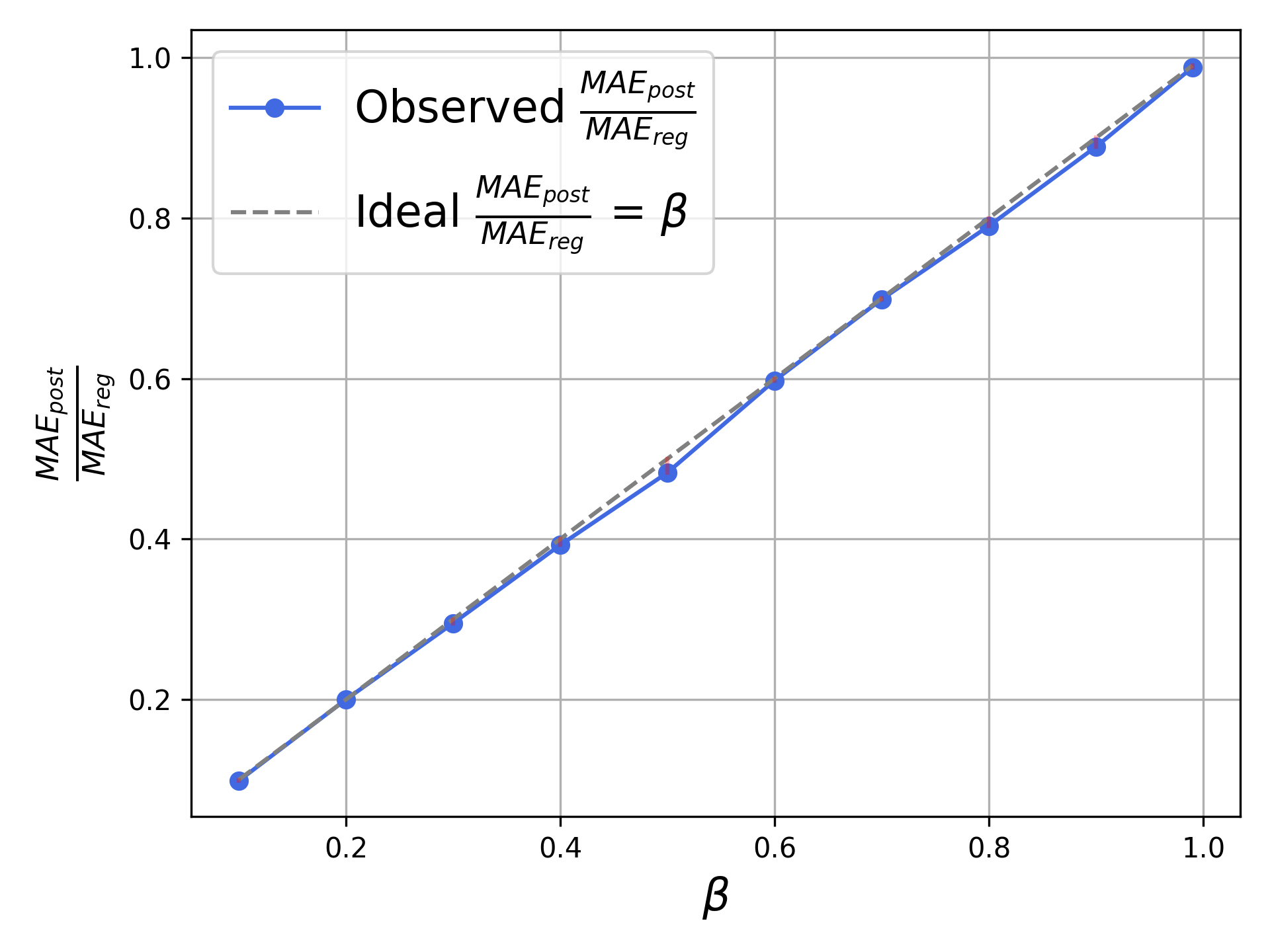}
  \caption{Empirical validation of the theoretical MAE reduction bound under ideal conditions. For each target improvement factor $\beta$, we compute the required ranker variance $\sigma^2_\text{rank}$ using the right-side inequality in Implication~\ref{eq:beta_bound}. We simulate predictions from an oracle regressor ($\hat{y}_0^{\text{reg}} \sim \mathcal{N}(y_0, 1)$) and an oracle ranker ($\hat{y}_0^\text{rank} \sim \mathcal{N}(y_0, \sigma^2_\text{rank})$), and compute the fused estimate. The observed post-refinement MAE ratios match the ideal $\beta$ values, demonstrating the correctness of the fusion rule under Gaussian distribution assumptions.}
\label{fig:oracle-beta}

\end{figure}

To validate Implication \ref{eq:beta_bound}, we simulate unbiased oracle predictions with known variances.  
For target factors $\beta\in\{0.10,0.20,\dots,0.90,0.99\}$, the required ranker variance $\sigma^{2}_{\text{rank}}$ is obtained from the right-side inequality of Implication~\ref{eq:beta_bound}.  
Synthetic estimates are drawn as  
$\hat{y}_0^{\text{reg}}\sim\mathcal{N}(y_0,1)$ and  
$\hat{y}_0^{\text{rank}}\sim\mathcal{N}(y_0,\sigma^{2}_{\text{rank}})$.  
The fused prediction $\hat{y}_0^{*}$ follows Equation~\ref{eq:ivw}.  
Figure \ref{fig:oracle-beta} confirms that empirical MAE ratios match the prescribed targets.

\subsection{Effect of Ranker Accuracy and Reference Sample Size on Molecular-Property Prediction with \RankRefine}
\label{sec:sweep_mpp}
\begin{figure}[htbp]
    \centering
    \begin{subfigure}[b]{0.325\textwidth}
        \includegraphics[width=\textwidth]{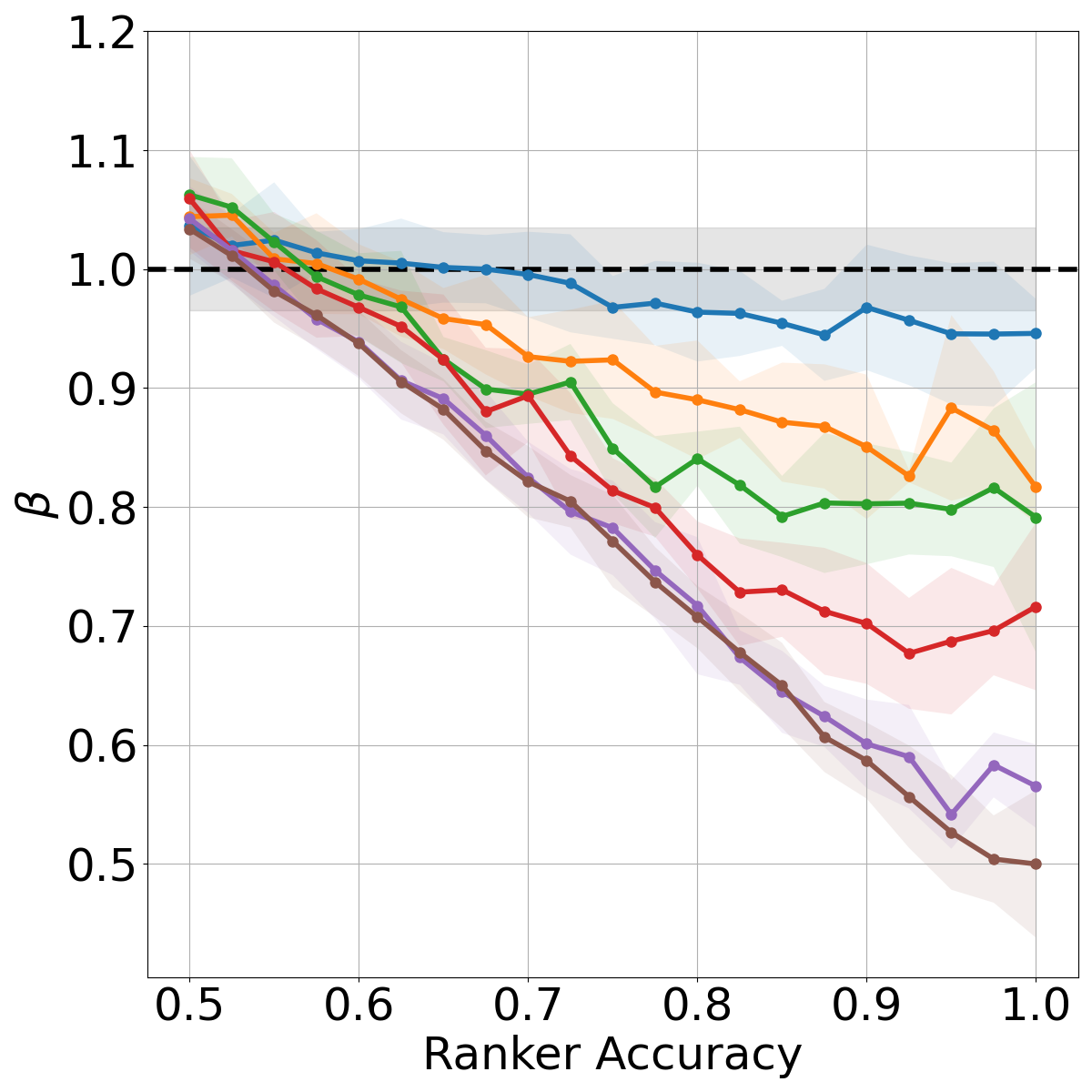}
        \caption{Caco2}
    \end{subfigure}
    \hfill
    \begin{subfigure}[b]{0.325\textwidth}
        \includegraphics[width=\textwidth]{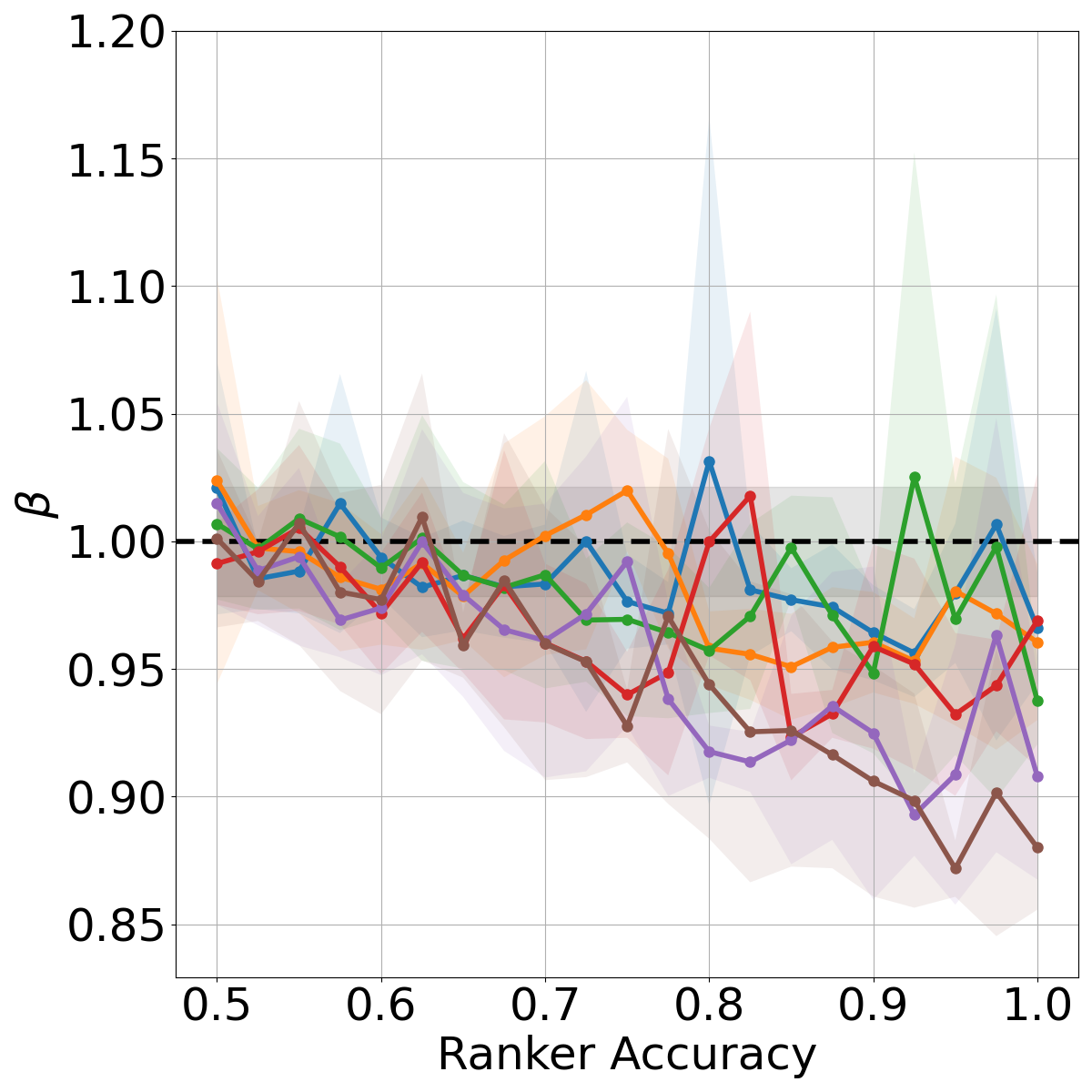}
        \caption{Clearance Microsome}
    \end{subfigure}
    \hfill
    \begin{subfigure}[b]{0.325\textwidth}
        \includegraphics[width=\textwidth]{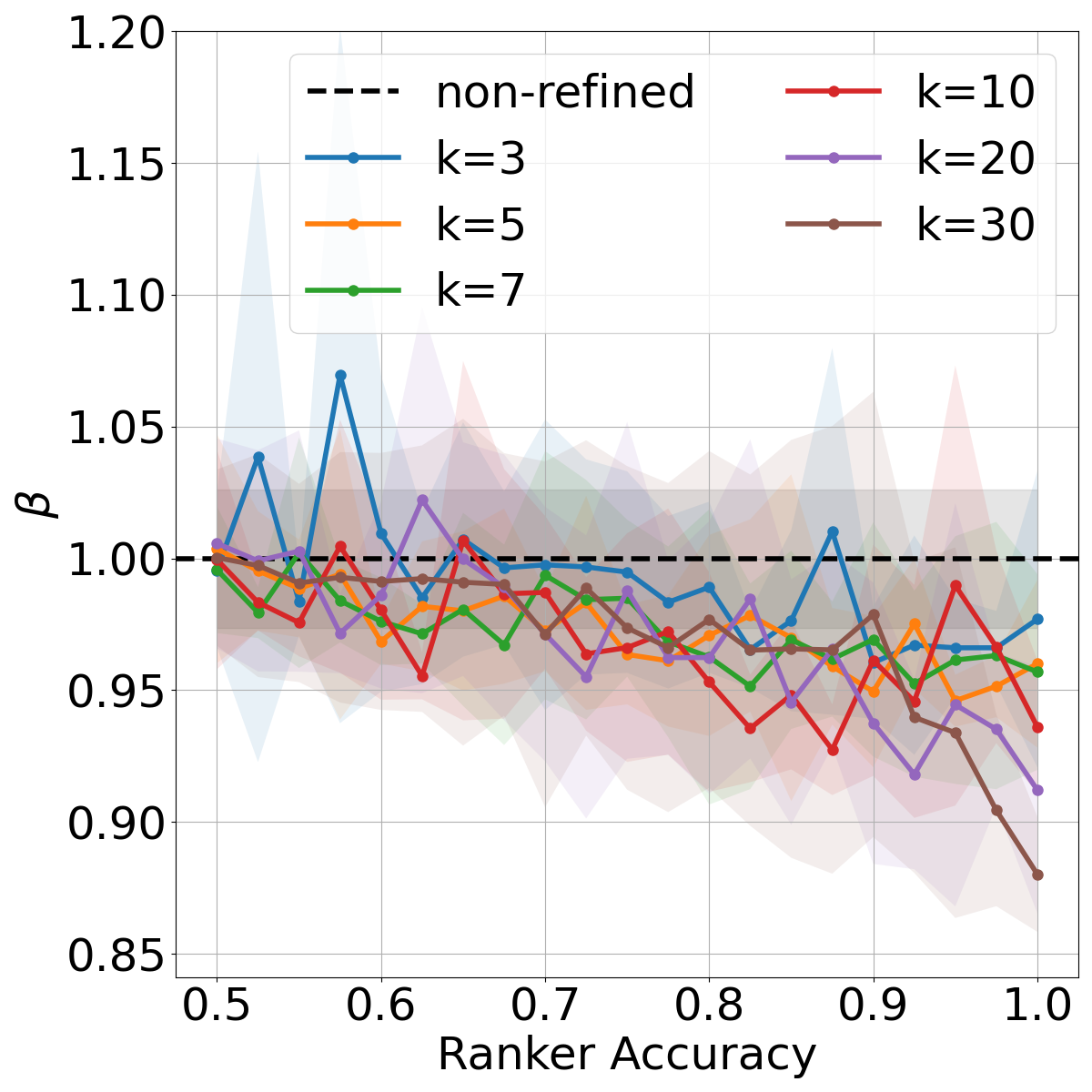}
        \caption{Clearance Hepatocyte}
    \end{subfigure}

    \vspace{0.5cm}

    \begin{subfigure}[b]{0.325\textwidth}
        \includegraphics[width=\textwidth]{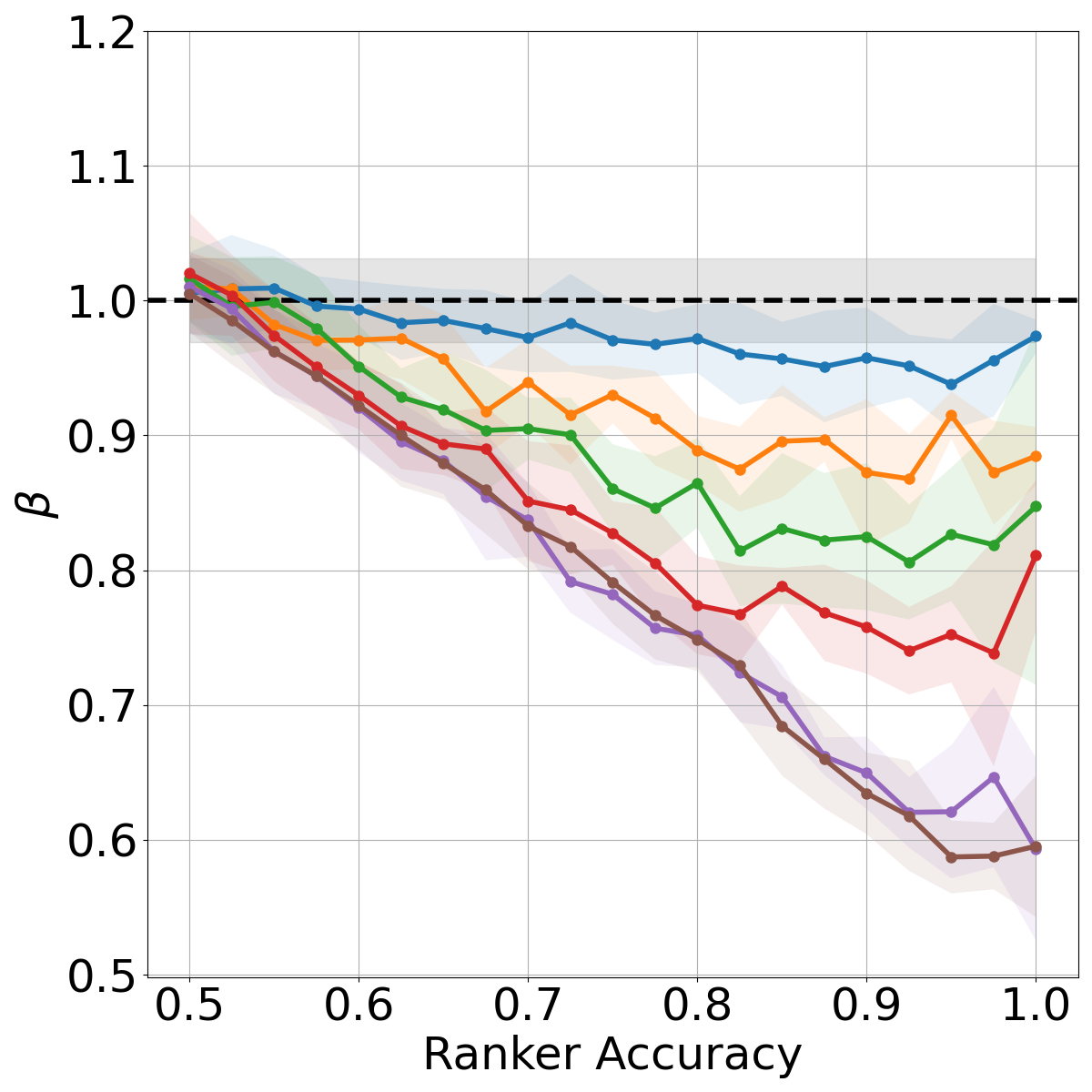}
        \caption{Half Life}
    \end{subfigure}
    \hfill
    \begin{subfigure}[b]{0.325\textwidth}
        \includegraphics[width=\textwidth]{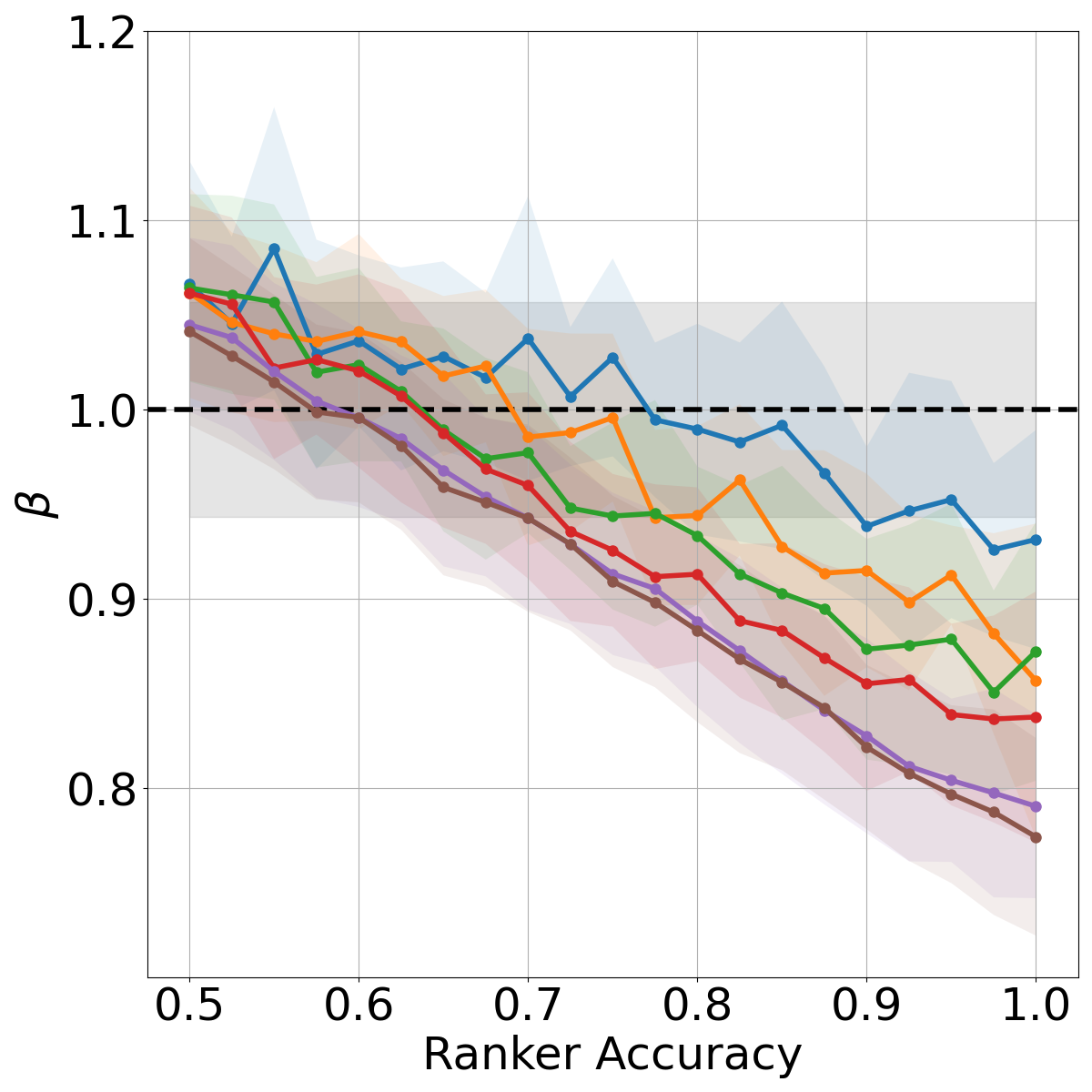}
        \caption{FreeSolv}
    \end{subfigure}
    \hfill
    \begin{subfigure}[b]{0.325\textwidth}
        \includegraphics[width=\textwidth]{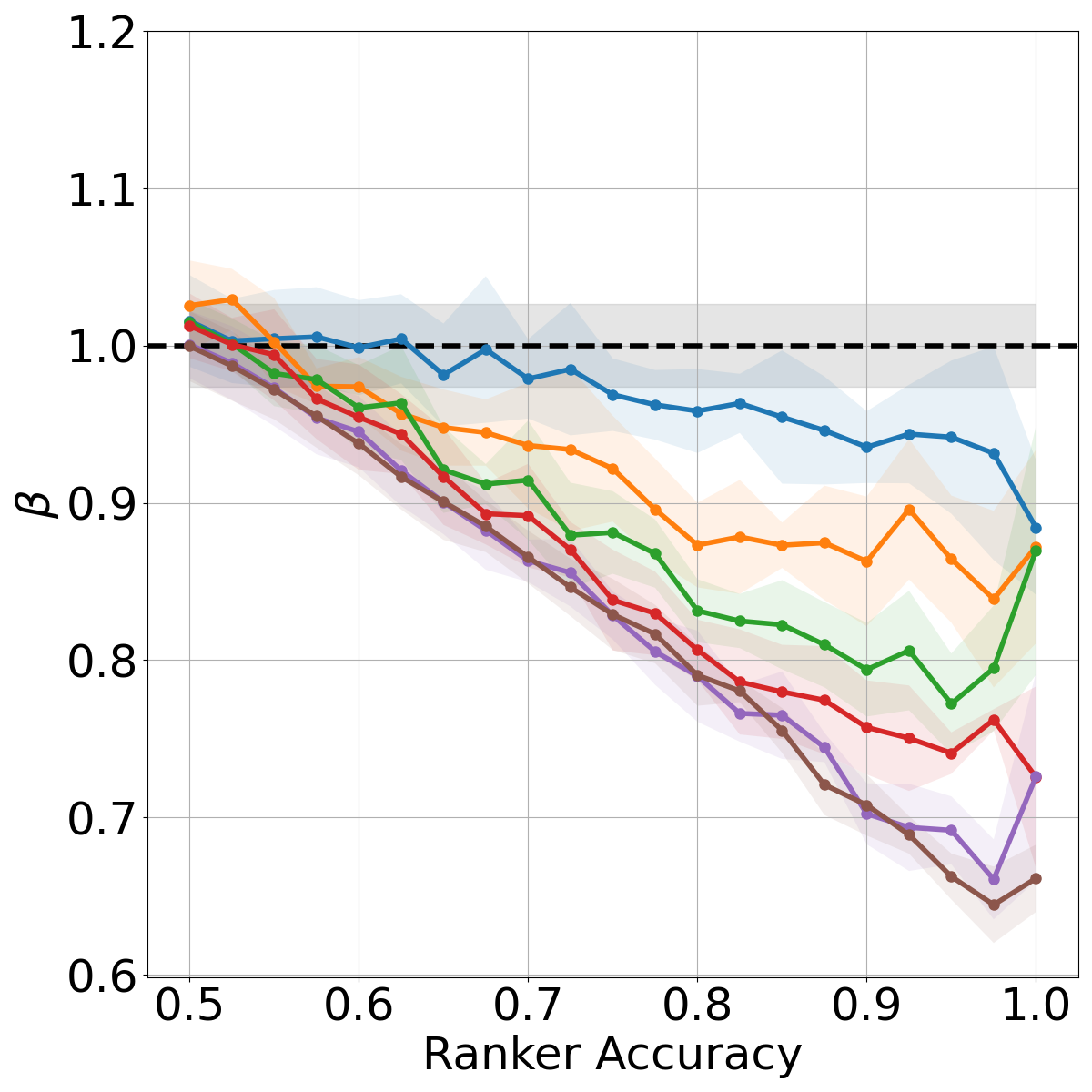}
        \caption{Lipophilicity}
    \end{subfigure}

    \vspace{0.5cm}

    \begin{subfigure}[b]{0.325\textwidth}
        \includegraphics[width=\textwidth]{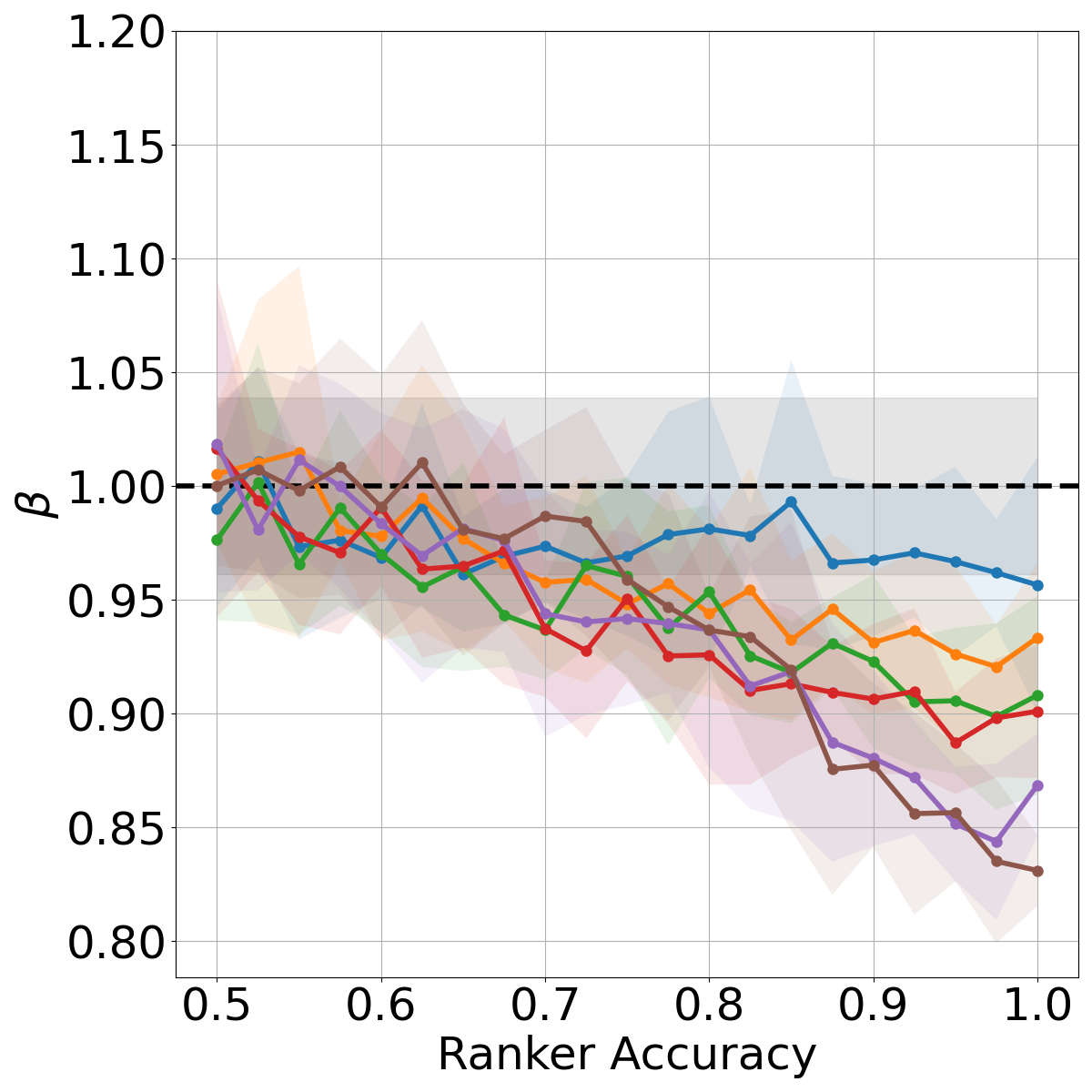}
        \caption{PPBR}
    \end{subfigure}
    \hfill
    \begin{subfigure}[b]{0.325\textwidth}
        \includegraphics[width=\textwidth]{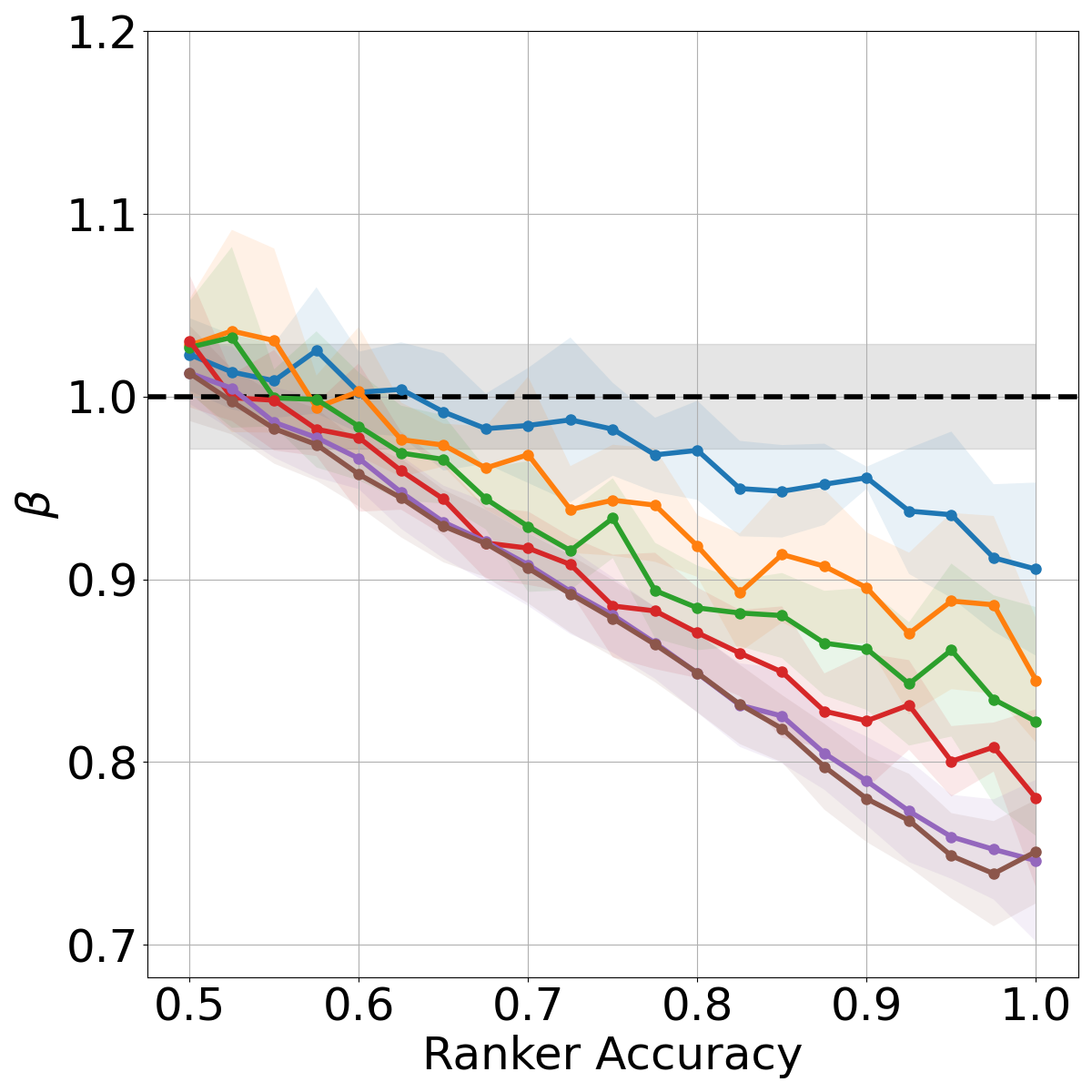}
        \caption{Solubility}
    \end{subfigure}
    \hfill
    \begin{subfigure}[b]{0.325\textwidth}
        \includegraphics[width=\textwidth]{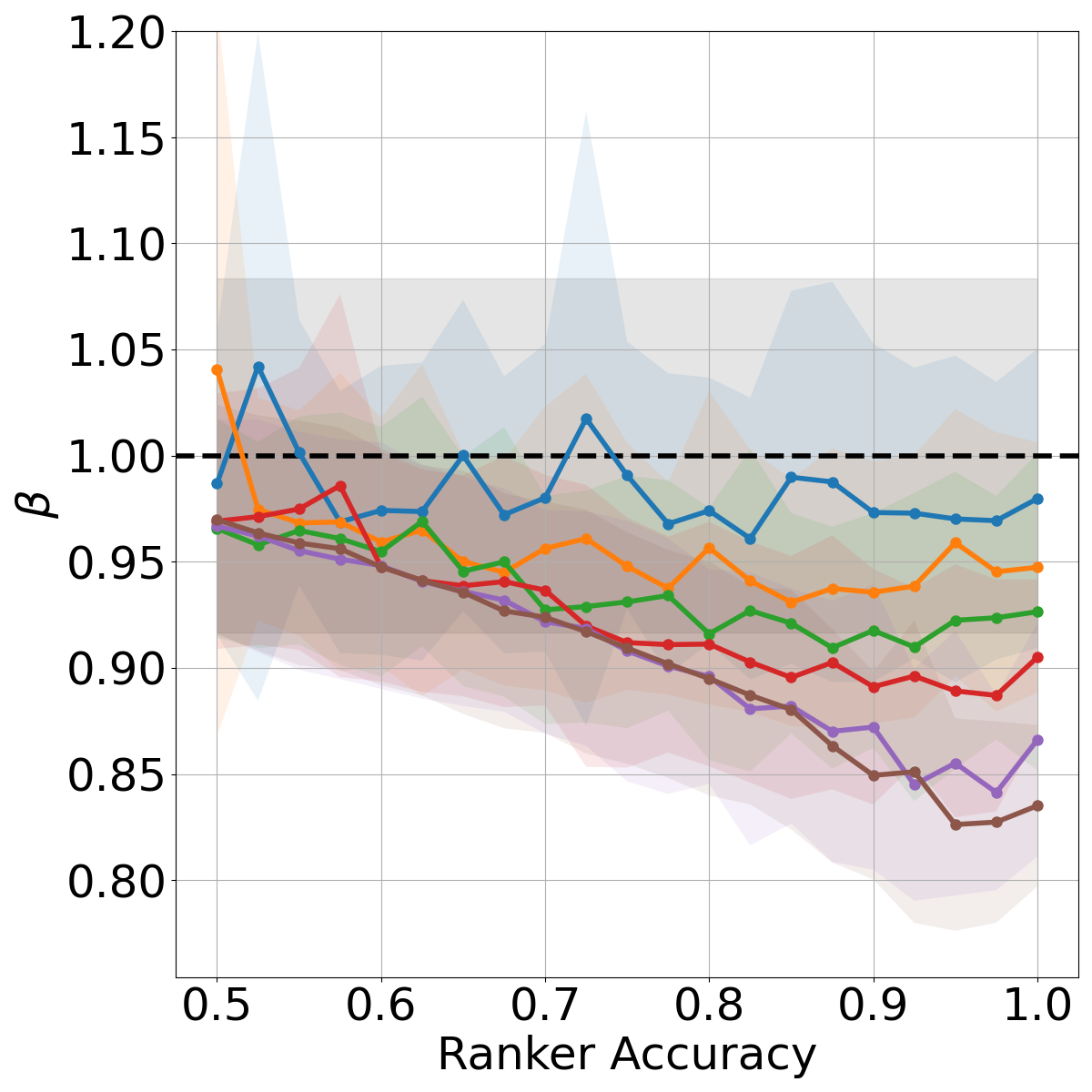}
        \caption{VDss}
    \end{subfigure}

    \caption{Performance of \RankRefine\ on molecular property prediction datasets under varying oracle ranker accuracy and number of reference comparisons. Each plot shows the normalized error $\beta = \frac{\text{MAE}_{\text{post}}}{\text{MAE}_{\text{reg}}} $ as a function of ranker accuracy, averaged over 5 random splits. $\beta < 1$ indicates improvements in regression performance over the base regressor. $k$ is the number of pairwise comparisons for each test molecule. Shaded regions indicate standard deviation. Dashed gray line shows baseline MAE with no refinement. Across most configurations, \RankRefine\ improves regression performance when using a ranker with accuracy as low as 0.55. Increasing $k$ tends to lower $\beta$, but typically the benefits start to diminish beyond $k=20$.}
    \label{fig:mpp_ranker_sweep}
\end{figure}

Using the nine TDC datasets, we vary oracle ranker accuracy and the number $k$ of reference comparisons per query.  
Figure \ref{fig:mpp_ranker_sweep} plots $\beta$, the post refinement MAE error divided by the regression error, versus pairwise ranker accuracy.  
\RankRefine\ consistently lowers error, even with accuracy $\approx0.55$ and $k=10$.  
Increasing the number of pairwise comparisons $k$ tend to further lower $\beta$. We observe that
$k = 20$ is optimal, as the difference between $k=20$ and $k = 30$ is often minimal.

For very accurate rankers ($>0.95$) a slight uptick in $\beta$ arises from over-confident curvature estimates and extreme solutions when the query is an outlier.
For the first source, recall that $\sigma^2_\text{rank}$ is approximated by the second derivative of the negative log-likelihood.
This curvature becomes smaller as the ranker accuracy approaches perfection, leading to over-confident estimates.
For the second source, consider the case when the query is truly the smallest or largest value among the reference set.
This is especially common when $k$ is small and ranker is accurate.
The optimizer may push $\hat{y}^{*\text{rank}}_0$ toward extreme values that still satisfy pairwise ranking constraints, but overshoot the true property value $y_0$.

\begin{figure}[htbp]
    \centering
    \begin{subfigure}[b]{0.325\textwidth}
        \includegraphics[width=\textwidth]{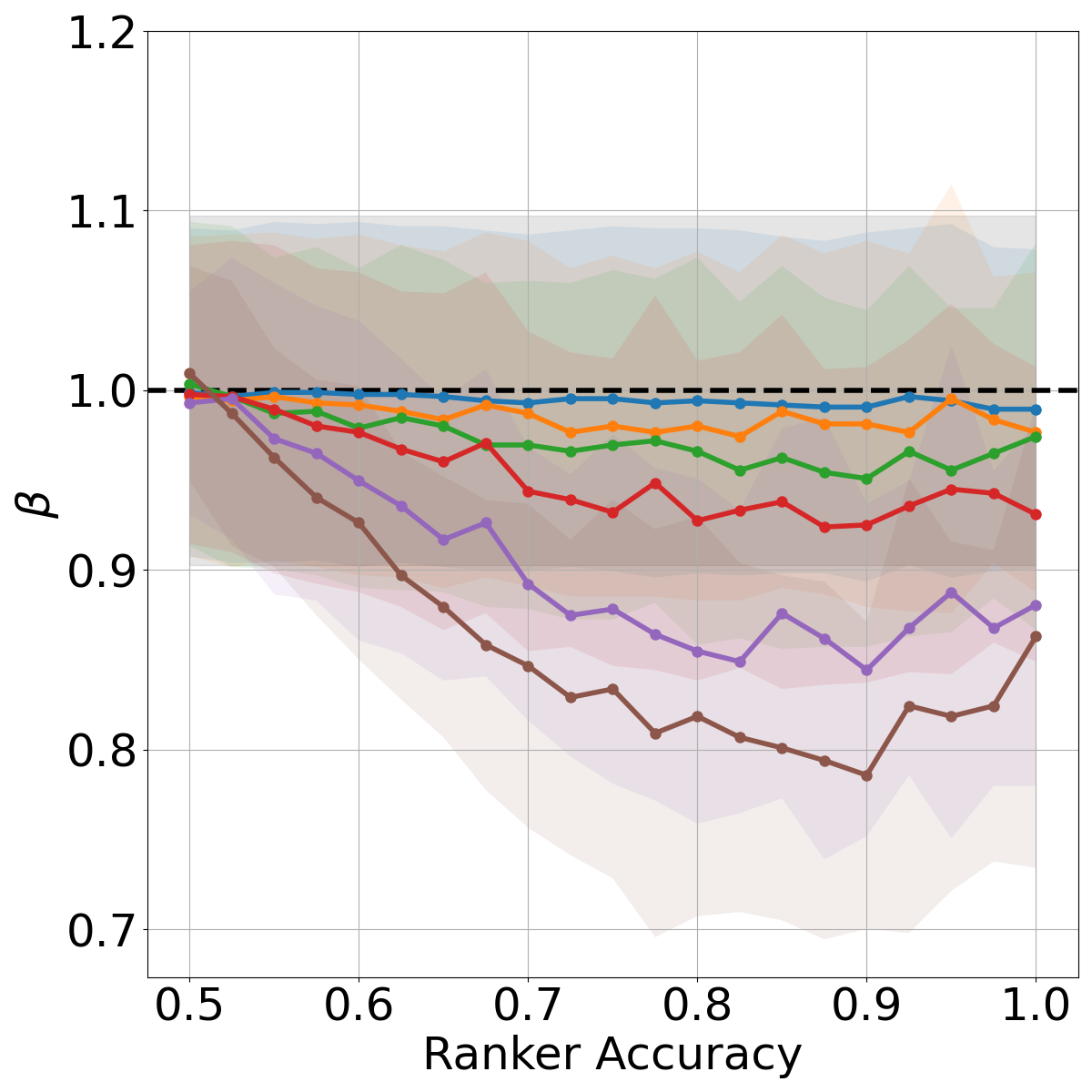}
        \caption{International Education Cost}
    \end{subfigure}
    \hfill
    \begin{subfigure}[b]{0.325\textwidth}
        \includegraphics[width=\textwidth]{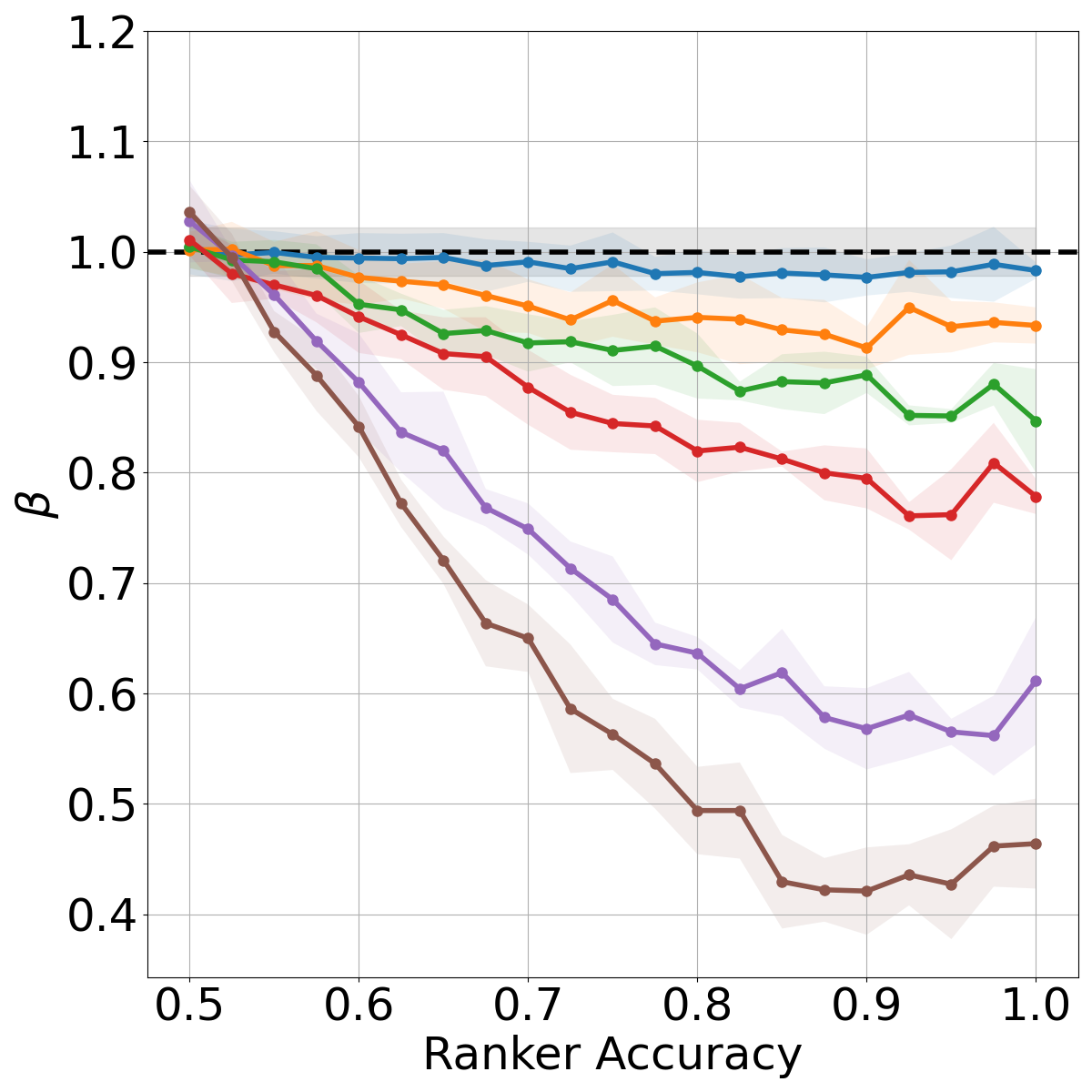}
        \caption{Smart Farming}
    \end{subfigure}
    \hfill
    \begin{subfigure}[b]{0.325\textwidth}
        \includegraphics[width=\textwidth]{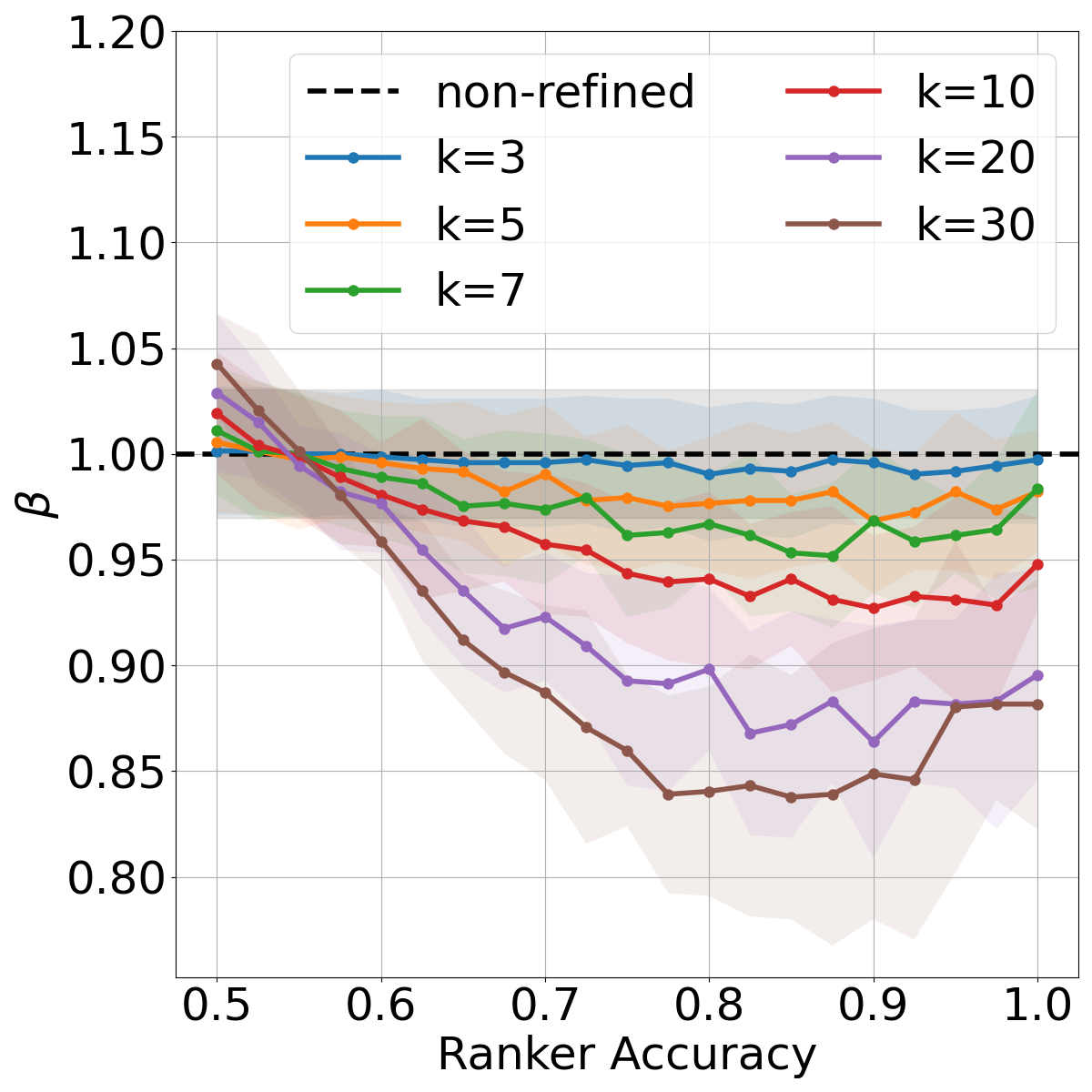}
        \caption{Student Performance}
    \end{subfigure}
    \caption{Performance of \RankRefine\ on tabular datasets under varying oracle ranker accuracy and number of reference comparisons. Across most configurations, \RankRefine\ improves regression performance when using a ranker with accuracy as low as 0.55.}
    \label{fig:sweep_kaggle}
\end{figure}

\subsection{Cross-Domain Generalization}
Applying the same protocol to the three tabular datasets yields Figure \ref{fig:sweep_kaggle}, whose reductions mirror those in chemistry datasets.
The results show that \RankRefine\ generalizes beyond molecular tasks.



\subsection{Comparison with Other Baselines}
\begin{figure}
    \centering
    \begin{subfigure}[b]{0.49\textwidth}
        \includegraphics[width=\textwidth]{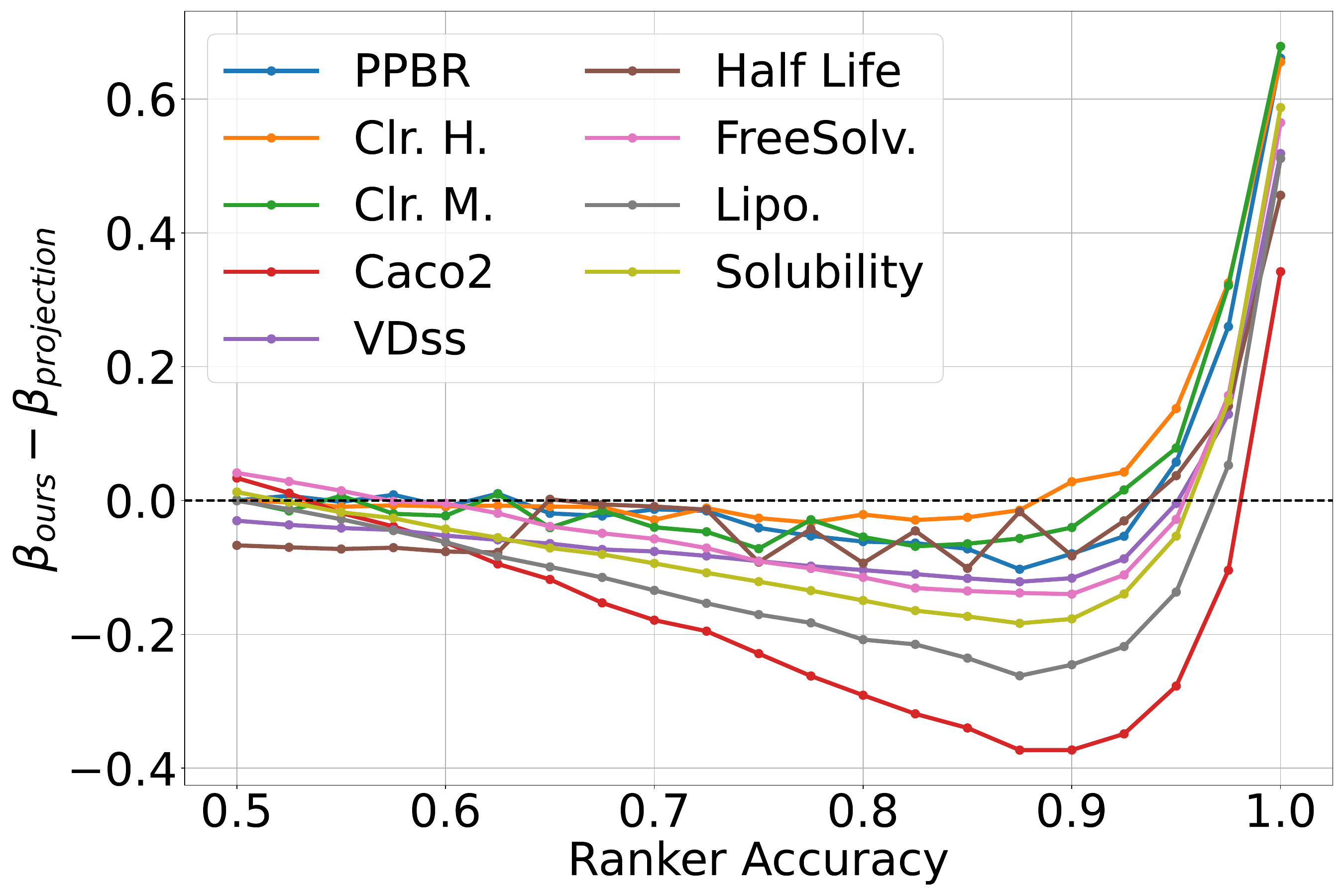}
        \caption{}
    \end{subfigure}
    \hfill
    \begin{subfigure}[b]{0.49\textwidth}
        \includegraphics[width=\textwidth]{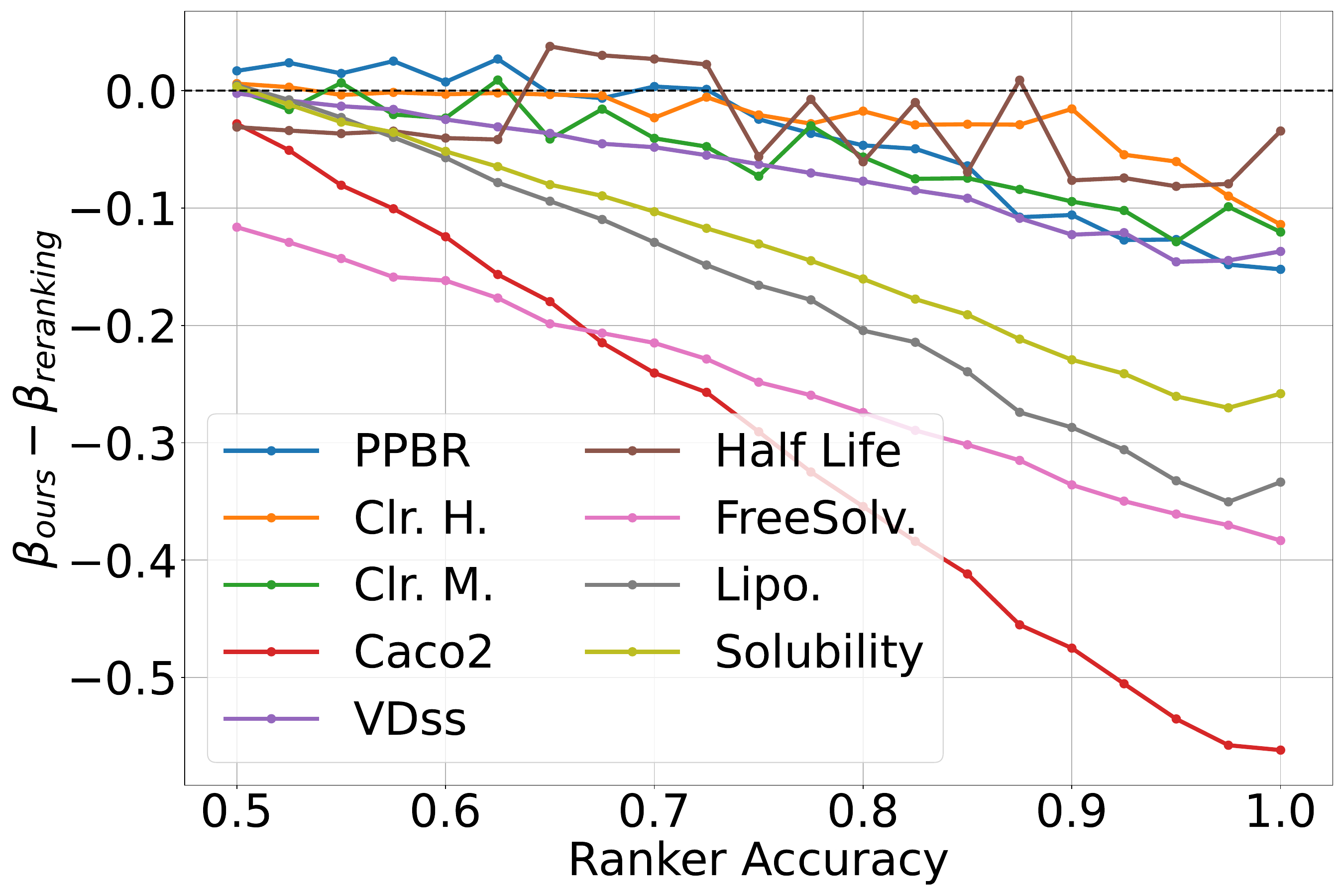}
        \caption{}
    \end{subfigure}
    \caption{(a) Comparison between \RankRefine\ the constrained optimization (projection)-based refinement method of \citet{yan2024consolidating_projection} on molecular property prediction tasks. We report the difference in normalized error, $\beta_\text{ours} - \beta_\text{projection}$. A negative value indicates better performance by \RankRefine. Each curve corresponds to a dataset. \RankRefine \ generally outperforms the baseline when the ranker accuracy is between 0.5 and 0.95. (b) We also compare our method to a related post hoc regression improvement method, regression by re-ranking \citep{gonccalves2023regression_regressionbyreranking}.}
    \label{fig:sweep_vs_baseline}
\end{figure}

We compare against two post hoc refinement methods: constrained optimization \citep{yan2024consolidating_projection} and regression by re-ranking (RbR) \citep{gonccalves2023regression_regressionbyreranking}.
We reimplemented the two baselines from scratch, as no publicly available code was provided.
We use  $k=30$ comparisons.  
Figure \ref{fig:sweep_vs_baseline} displays $\beta_{\text{ours}}-\beta_{\text{projection}}$ and $\beta_{\text{ours}}-\beta_{\text{reranking}}$; negative values favour \RankRefine.
Our approach generally outperforms RbR in all datasets.
Against the projection-based baseline, \RankRefine\ excels when ranker accuracy lies between 0.50 and 0.95; projection dominates only when the ranker is nearly perfect.
This is because, when all pairwise rankings are correct, the projection method strictly enforces the constraints without increasing the absolute error.
In such cases, the MAE will either remain unchanged or improved, with the improvements depend primarily on the distribution and resolution of the known labels.
Yet, such near-perfect rankers are rare in real-world scenarios.
Thus, the robust performance of \RankRefine\ with moderately-accurate rankers offers more practicality.
More comparisons with projection-based refinement can be found in the supplementary materials.

\subsection{Few-Shot Molecular-Property Prediction with LLMs}

\begin{table}[t]
    \centering
    \caption{PRA and resulting \RankRefine\ performance using ChatGPT-4o as the ranker on six datasets from the TDC ADME benchmark. For each test molecule, 20 training molecules are randomly sampled to form pairwise comparisons. Values are averaged over 3 random splits. Despite moderate pairwise accuracy ($\sim$0.60–0.69), \RankRefine\ consistently improves regression performance, demonstrating the viability of LLM-based rankers in low-data  domains.\\}
    \begin{tabular}{rccc}
    \toprule
        Dataset & Lipophilicity & Solubility & VDss \\
    \cmidrule(lr){1-4}
        Pairwise Ranking Accuracy & 0.622	±	0.008 & 0.693	±	0.035 & 0.605	±	0.010 \\
        $\beta$ & 0.957 ± 0.012 & 0.934 ± 0.048 & 0.895 ± 0.053 \\
    \toprule
        Dataset & Caco2 & Half Life & FreeSolv \\
    \cmidrule(lr){1-4}
        Pairwise Ranking Accuracy & 0.660	±	0.013 & 0.602	±	0.014 & 0.681	±	0.050  \\
        $\beta$ & 0.970 ± 0.027 & 0.971 ± 0.005 & 0.937 ± 0.012 \\
        
    \bottomrule \\
    \end{tabular}
    \label{tab:chatgpt_result}
\end{table}
To evaluate the practical applicability of \RankRefine\ in realistic few-shot settings, we replace the oracle ranker with ChatGPT-4o \citep{chatgpt4o}, a large language model (LLM). Note that we can still evaluate the ranking accuracy of the LLM through oracle for the purpose of analysis. 
For each test molecule, we randomly sample 20 reference molecules from the training set and ask ChatGPT-4o to perform pairwise comparisons (e.g., “Which molecule is likely to have higher solubility?”).
The model is queried using textual molecular descriptions, SMILES \citep{weininger1988smiles}. 
The details of the prompt is provided in the supplementary materials.

Table \ref{tab:chatgpt_result} summarizes ChatGPT-4o ranking accuracy and the resulting $\beta$ on six TDC datasets.  
Even moderate accuracies ($\approx$0.62–0.69) yield tangible MAE gains, underscoring both the robustness and practicality of our method.

One natural concern when employing a pretrained LLM as a ranker is whether it simply memorizes from training data which item (e.g., a molecule) has a higher or lower property value (e.g., solubility) compared to others.
To test this, we conducted an additional evaluation on a private compound-activity dataset consisting of 75 compounds with IC50 labels measured against a specific target protein complex.
This dataset is unpublished, and therefore, not part of any LLM training corpora, ruling out direct memorization.
On this dataset, ChatGPT-4o achieves 60.14\% pairwise ranking accuracy, indicating generalization capability and not simple memorization.

\subsection{Human Age Estimation with Pairwise Self-Correction}
\begin{table}[t]
    \centering
    \caption{Performance of \RankRefine\ when using human participants as both direct estimators and pairwise rankers on a facial age estimation task. The base error ($\text{MAE}_{\text{reg}}$) corresponds to the raw human estimate error. Each test face is compared to 15 reference faces via human-judged pairwise comparisons. \RankRefine\ improves prediction accuracy by refining each estimate based on human-provided rankings.\\}
    
    \begin{tabular}{ccc}
    \toprule
        $\text{MAE}_\text{reg}$ & Pairwise Ranking Acc. & $\beta$ \\
    \cmidrule(lr){1-3}
         6.343 ± 0.610 & 0.759 ± 0.052 & 0.954 ± 0.046 \\
    \bottomrule\\
    \end{tabular}
    \label{tab:human_ranker}
\end{table}
\RankRefine\ can potentially be used in an interactive, human-in-the-loop scenarios, where domain experts act as the pairwise rankers.
As a proof of concept, we conduct a user study on the age estimation task.
Details about the user study is available in the supplementary materials.

Table \ref{tab:human_ranker} reports a UTKFace user study where 6 participants are asked to: (1) estimate the age of 17 individuals from their corresponding pictures, and (2) perform pairwise judgments to 15 individuals in the reference set.  
Pairwise judgments raise ranking accuracy to 0.76 and cut MAE by $\approx$5 \%, illustrating \RankRefine’s value for expert self-correction.

\subsection{Effects of Noisy Variance Estimates}

\begin{figure}[ht]
    \centering
     \begin{subfigure}[b]{0.48\textwidth}
        \includegraphics[width=\textwidth]{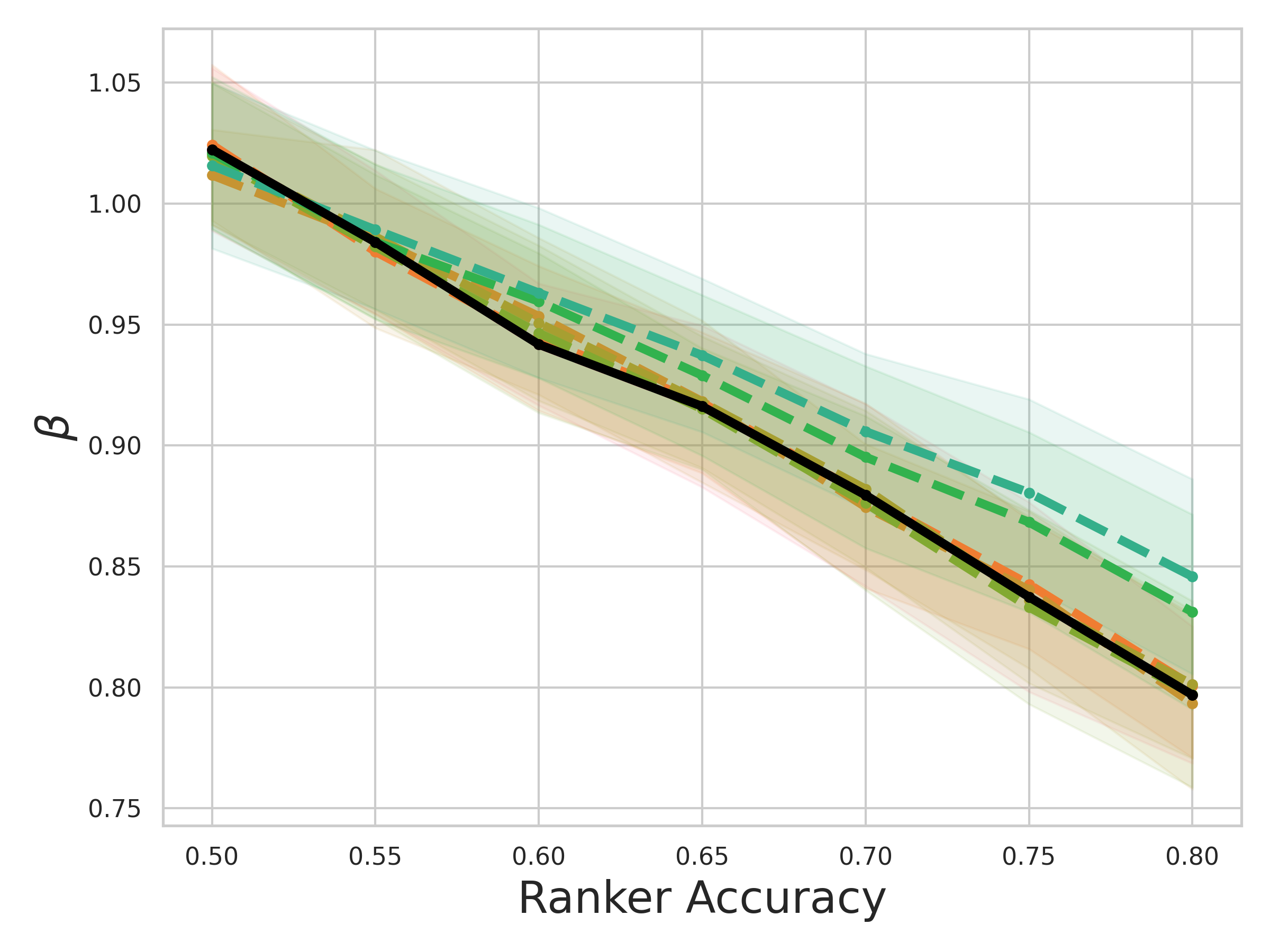}
        \caption{k = 30}
    \end{subfigure}
    \hfill
     \begin{subfigure}[b]{0.48\textwidth}
        \includegraphics[width=\textwidth]{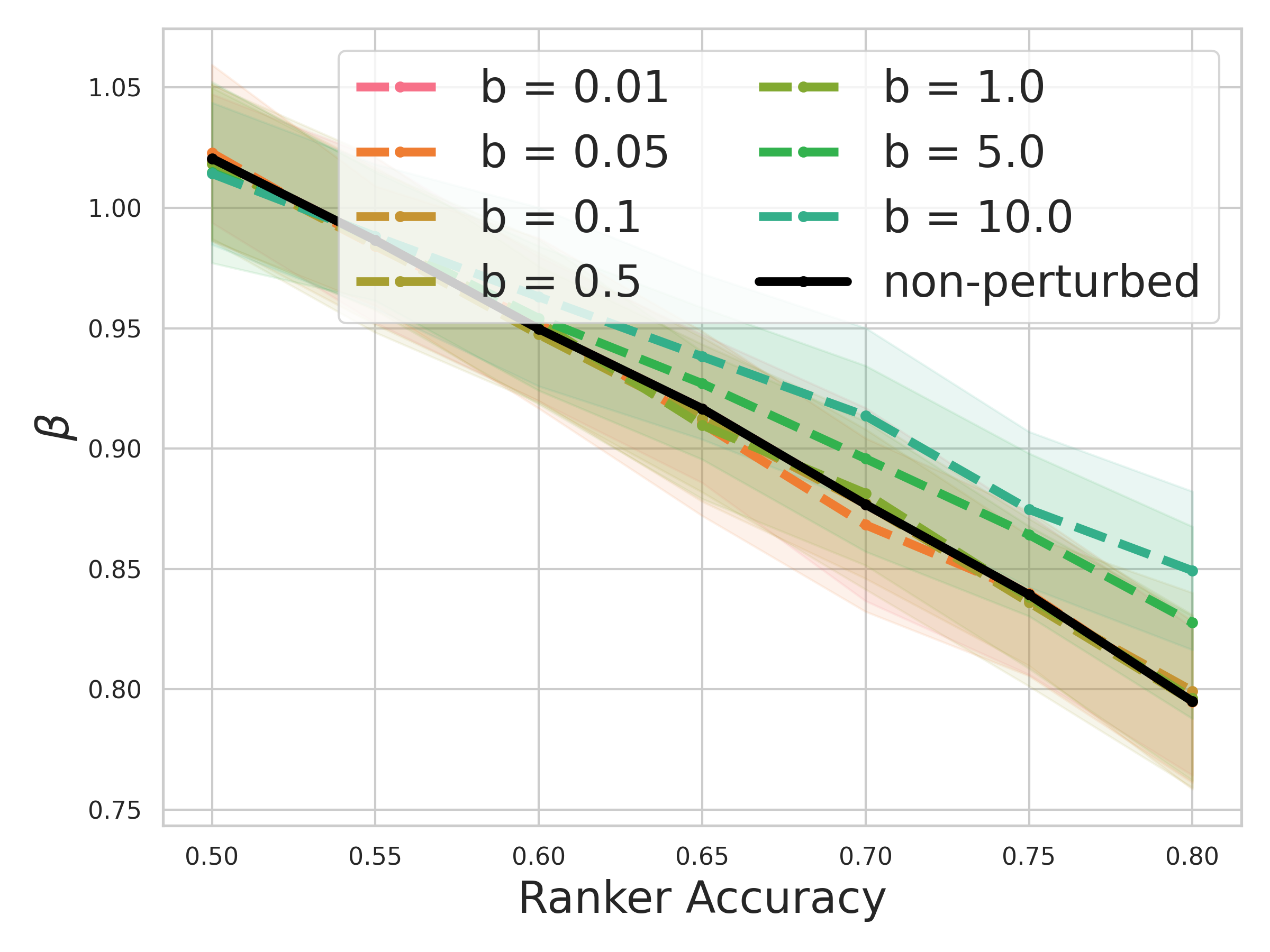}
        \caption{k = 40}
    \end{subfigure}
    \hfill
    \caption{Impact of noisy ranker variance estimate on the regression performance in Solubility dataset. Uniform noise from [-b, +b] is added to the estimated ranker variance. Performance starts to degrade when $b\geq5$, more than 3 times the standard deviation of the ranker variance estimate.}
    \label{fig:noisy_var}
\end{figure}

We study the effect of noise in the ranker variance estimate on the Solubility dataset using $k = 30$ and $k = 40$.
We add uniform perturbations sampled from [-b, +b] to the estimated ranker variance before fusion.
As shown in Figure \ref{fig:noisy_var}, performance degradation becomes noticeable when $b\geq5$, which is more than three times the standard deviation of the ranker variance estimates.
For reference, the mean and standard deviation of the estimated ranker variance are 2.358 ± 1.079.

\subsection{Effects of Biased Regressor}
We systematically inject constant bias into an otherwise unbiased regressor.
Specifically, we uniformly draw ground-truth labels $y_0 \sim U(0, 1)$, and set predictions by injecting zero-mean Gaussian noise $\hat{y}_0^{reg} = y_0 + \epsilon$, such that $\text{MAE} \approx 1\text{SD}$ where SD is the standard deviation of the ground-truth labels.
We then add a constant offset, $B$, increasing in magnitude from 0\% to 70\% of the label standard deviation (SD).
We report the refined-to-original MAE ratio, $\beta$, for varying ranker accuracies.
Table \ref{tab:ablation_biasRegressor} in Appendix D shows that $\beta$ worsens as bias increases, but remains $<$1 up to $B$ = 60\% SD, indicating that RankRefine remains effective even under substantial regressor bias. 

\subsection{Effects of Biased Sampling}
We study the effects of non-uniform sampling of references, in particular, when the references are clustered.
Similar to the biased regressor experiment, we generate synthetic predictions with ground-truth label range $[0,1]$.
We then restrict the reference labels to increasingly narrow subranges of the ground-truth label range, centered around the midpoint of the range.
The reference set bias, $RB = x\%$, means that the reference labels cover only $(100-x)\%$ of the full ground-truth range.
For example, an $RB$ = 10\% means the reference labels range is [0.05, 0.95].
Table \ref{tab:ablation_biasSampling} in Appendix E suggests that RankRefine performance is best when the reference set spans the full ground-truth labels' range (i.e., $RB$ = 0\%), but clustered sets still yield substantial gains.
For example, at RB = 90\% and ranker accuracy = 60\%, $\beta$ = 0.8842, an 11.58\% improvement.

Furthermore, we evaluate biased sampling in two challenging settings: (i) \textit{disjoint}: reference labels are in [-1,0] or [1,2] while queries are in [0,1]; and (ii) \textit{partial overlap}: reference labels are in [-0.5, 0.5] or [0.5, 1.5], while queries are in [0,1]. 
This simulates a distribution shift, where the distribution of the reference labels is different from the distribution of the query labels.
Despite the challenging setups, RankRefine improves the base regressor when the ranker is moderately accurate (i.e., $\geq$65\% in the disjoint case, $\geq$55\% in the partial overlap case). 
More details are available in Table \ref{tab:shifted_sampling} in Appendix E.

\section{Limitations}
While \RankRefine\ demonstrates strong empirical performance, it has a number of limitations: 
\begin{itemize}
    \item Our theoretical analysis assumes that both the regressor and rank-based estimates have unbiased, Gaussian-distributed errors and are independent. Despite the empirical robustness of our method, its assumptions may not hold in practice, particularly in the presence of heavy-tailed or skewed noise. 
    \item \RankRefine\ depends on well-calibrated uncertainty estimates from the regressor and a trustworthy estimate of ranker variance (via Equation~\ref{eq:sigma_rank_estimate}). 
    Miscalibration in either component can reduce or even negate the performance gains.
    \item Our oracle experiments assume uniformly random ranking errors; however, real-world rankers may exhibit systematic biases, for example consistently failing on extreme values which can skew the fusion process. 
    \item \RankRefine\ uses the Bradley–Terry model by treating true property values as proxies for latent scores, which differs from its original use where latent scores are learned to explain probabilistic pairwise outcomes. While practical, this introduces a modeling mismatch. A possible solution is to learn a mapping from a standard Bradley-Terry latent scores to labels on a holdout set and apply it during inference

\end{itemize}

\section{Future Work}
Future work can extend \RankRefine\ in several promising directions. 
One avenue is to replace the Bradley-Terry likelihood with richer stochastic-transitivity models that better capture structured or systematic errors in the ranker. 
Another important direction involves incorporating Bayesian or conformal calibration layers to automatically correct misestimated variances and improve reliability. 
Beyond scalar predictions, \RankRefine\ could potentially be extended to handle multivariate or structured targets, such as full pharmacokinetic profiles. 
Finally, leveraging rationale from language models could improve the interpretability of the ranking process and facilitate richer forms of expert feedback in decision-critical domains.

\section{Conclusion}
We introduced \RankRefine, a plug-and-play, post hoc framework that injects pairwise-ranking signals into any uncertainty-aware regressor.
Theoretically, we proved that inverse-variance fusion with a rank-based estimator lowers the expected mean absolute error (MAE) whenever the ranker variance is finite.
Empirically, oracle simulations generally follows the theory, and experiments on nine molecular-property benchmarks plus three diverse tabular tasks showed consistent improvements even with ranker accuracies of around $55\%$ and as few as 10 comparisons.
ChatGPT-4o rankings yielded measurable MAE drops on six ADME datasets, and a user study demonstrated similar self-correction for human age estimation. 
Compared with a projection baseline, \RankRefine\ prevailed whenever ranker accuracy lay in the realistic $0.50$–$0.95$ range, underscoring its robustness and broad applicability.

\bibliography{neurips}


\newpage
\appendix
\section{Extended Proof}
\subsection{\RankRefine\ is a minimum-variance unbiased estimator}
Suppose that we have two independent unbiased estimates of the same parameter $y_0$,
\[
\hat{y}_0^\text{reg} \sim \mathcal{N}(y_0, \sigma^2_\text{reg}), \qquad \hat{y}_0^\text{rank} \sim \mathcal{N}(y_0, \sigma^2_\text{rank}),
\]
and we want to form a linear combination of the two estimates,
\begin{equation}
    \hat{y}_0^* = w \hat{y}_0^\text{reg} + (1-w) \hat{y}_0^\text{rank},
    \label{eq:minimum_variance_linear_combination}
\end{equation}
our goal is to choose $w \in [0,1]$ such that the variance is minimized.

For two independent random variables $X \text{ and } Y$, $\text{Var}(aX + bY) = a^2 X + b^2 Y$.
The post-fusion variance of the unbiased estimates is then,
\begin{equation}
    \sigma^2_\text{post} = w^2\sigma^2_\text{reg} + (1-w)^2 \sigma^2_\text{rank}.
    \label{eq:minimum_variance_var_post}
\end{equation}
To minimize $\sigma^2_\text{post}$ with respect to $w$, we take the derivative and solve for,
\begin{equation}
\begin{split}
    \frac{d}{dw}\bigl(w^2\sigma^2_\text{reg} + (1-w)^2\sigma^2_\text{rank} \bigr) &= 0,\\    
    2w\sigma^2_\text{reg} - 2(1-w)\sigma^2_\text{rank} &= 0\\
    2w\sigma^2_\text{reg} &= 2(1-w) \sigma^2_\text{rank}\\
    w\sigma^2_\text{reg} &= (1-w) \sigma^2_\text{rank}\\
    w\sigma^2_\text{reg} &= \sigma^2_\text{rank} - w\sigma^2_\text{rank}\\
    w\sigma^2_\text{reg} + w\sigma^2_\text{rank} &= \sigma^2_\text{rank}\\
    w(\sigma^2_\text{reg} + \sigma^2_\text{rank}) &= \sigma^2_\text{rank}. 
\end{split}
\end{equation}
Therefore,
\begin{equation}
    w = \frac{\sigma^2_\text{rank}}{\sigma^2_\text{reg} + \sigma^2_\text{rank}}, \qquad (1-w) = \frac{\sigma^2_\text{reg}}{\sigma^2_\text{reg} + \sigma^2_\text{rank}}.
    \label{eq:optimal_w}
\end{equation}
Substituting the weights in Equation \ref{eq:minimum_variance_var_post} with Equation \ref{eq:optimal_w}, we obtain the \RankRefine\ post-fusion variance in Equation \ref{eq:ivw},
\begin{equation}
\begin{split}
    \sigma^2_\text{post} &= \left(\frac{\sigma^2_\text{rank}}{\sigma^2_\text{reg} + \sigma^2_\text{rank}}\right)^2 \sigma^2_\text{reg} + \left(\frac{\sigma^2_\text{reg}}{\sigma^2_\text{reg} + \sigma^2_\text{rank}}\right)^2 \sigma^2_\text{rank}\\
    &= \frac{(\sigma^2_\text{rank})^2 \sigma^2_\text{reg} + (\sigma^2_\text{reg})^2 \sigma^2_\text{rank}}{(\sigma^2_\text{reg} + \sigma^2_\text{rank})^2}\\
    &= \frac{\sigma^2_\text{rank} \sigma^2_\text{reg} (\sigma^2_\text{rank} + \sigma^2_\text{reg})}{(\sigma^2_\text{reg} + \sigma^2_\text{rank})^2}\\
    &= \frac{\sigma^2_\text{rank} \sigma^2_\text{reg}}{\sigma^2_\text{rank} + \sigma^2_\text{reg}}\\
    &= \left( \frac{1}{\sigma^2_\text{reg}} + \frac{1}{\sigma^2_\text{rank}} \right)^{-1}
\end{split}    
\end{equation}

Similarly, substituting the weights in Equation \ref{eq:minimum_variance_linear_combination} with Equation \ref{eq:optimal_w}, we get the \RankRefine\ estimate in Equation \ref{eq:ivw},
\begin{equation}
\begin{split}
    \hat{y}_0^* &= \frac{\sigma^2_\text{rank}}{\sigma^2_\text{reg} + \sigma^2_\text{rank}}\ \hat{y}_0^\text{reg} + \frac{\sigma^2_\text{reg}}{\sigma^2_\text{reg} + \sigma^2_\text{rank}}\ \hat{y}_0^\text{rank}\\
    &= \frac{\sigma^2_\text{rank} \sigma^2_\text{reg}}{(\sigma^2_\text{reg} + \sigma^2_\text{rank}) \sigma^2_\text{reg}} \hat{y}_0^\text{reg}
+ \frac{\sigma^2_\text{reg} \sigma^2_\text{rank}}{(\sigma^2_\text{reg} + \sigma^2_\text{rank}) \sigma^2_\text{rank}} \hat{y}_0^\text{rank}\\
&= \frac{\sigma^2_\text{post}}{\sigma^2_\text{reg}} \hat{y}_0^\text{reg} + \frac{\sigma^2_\text{post}}{\sigma^2_\text{rank}} \hat{y}_0^\text{rank}\\
&= \sigma^2_\text{post} \left( \frac{\hat{y}_0^\text{reg}}{\sigma^2_\text{reg}} + \frac{\hat{y}_0^\text{rank}}{\sigma^2_\text{rank}} \right)
\end{split}
\end{equation}

\subsection{Analysis of \RankRefine}
Corollary \ref{cor:any_informative_ranker} is obtained from:
\begin{equation}
    \begin{split}
        \sigma^2_\text{post} < \sigma^2_\text{reg} &\iff \left(\frac{1}{\sigma^2_{\text{reg}}} + \frac{1}{\sigma^2_{\text{rank}}}\right)^{-1} < \sigma_{\text{reg}}^2,\\
        &\iff \frac{1}{\sigma^2_{\text{reg}}} + \frac{1}{\sigma^2_{\text{rank}}} > \frac{1}{\sigma_{\text{reg}}^2},\\
        &\iff \frac{1}{\sigma^2_\text{rank}} < 0,\\
        &\iff \sigma^2_\text{rank} < \infty.
    \end{split}
\end{equation}
For an unbiased Gaussian estimates $\hat{y}_0 \sim \mathcal{N}(y_0, \sigma^2)$, the expected error is,
\begin{equation}
    \begin{split}
        \mathbb{E}(|\hat{y}_0 - y_0|) &= \mathbb{E}(|y_0 + \epsilon - y_0|), \qquad \text{with } \epsilon\sim\mathcal{N}(0, \sigma^2)\\
        &= \mathbb{E}(|\epsilon|).
    \end{split}    
\end{equation}
Since $\epsilon$ is a Gaussian distribution, then $|\epsilon|$ follows the half-Gaussian distribution with an expected value equals to $\sqrt{2/\pi}\sigma$.
The subsequent $\beta$ bound in Equation \ref{eq:beta_bound} is obtained from:
\begin{equation}
    \begin{split}
        \text{MAE}_{\text{post}}\leq\beta\; \text{MAE}_{\text{reg}} &\iff \sigma_{\text{post}}\leq\beta\;\sigma_{\text{reg}},\\
        &\iff \sigma^2_{\text{post}}\leq\beta^2\;\sigma^2_{\text{reg}},\\
        &\iff \left(\frac{1}{\sigma^2_{\text{reg}}} + \frac{1}{\sigma^2_{\text{rank}}}\right)^{-1} \leq \beta^2\; \sigma_{\text{reg}}^2\\
        &\iff \frac{1}{\sigma^2_{\text{reg}}} + \frac{1}{\sigma^2_{\text{rank}}} \geq \frac{1}{\beta^2\;\sigma_{\text{reg}}^2}\\
        &\iff \frac{1}{\sigma^2_{\text{rank}}} \geq \frac{1 - \beta^2}{\beta^2\; \sigma_{\text{reg}}^2},\\
        &\iff \sigma_{\text{rank}}^{2}\leq
    \frac{\beta^{2}\;\sigma_{\text{reg}}^{2}}{1-\beta^{2}}
    \end{split}
\end{equation}

\section{Additional Experimental Details}
\subsection{Prompt for obtaining pairwise rankings from ChatGPT-4o}
We use the following prompt to query ChatGPT-4o to predict the pairwise rankings for the Lipophilicity dataset.

\begin{verbatim}
# Identity

You are an expert in chemistry and biology. Given a short description of a 
molecular property and two molecular SMILES, you can determine if Molecule A 
has greater property value than Molecule B or not.

# Instructions

* The list of pairwise molecules is given in two CSV files: "test_A.csv" and 
"test_B.csv". These are two different lists that you need to perform rank 
prediction on. The first line is the header.
* You cannot use external cheminformatics library to directly predict the property.
* You can design your own heuristics, comparing atom types, bonds, 
and other important information to make your predictions.
* Before you start designing the heuristics, briefly explain the property 
and what influence its values. Incorporate this prior knowledge into your heuristic.
* You are given 5 examples in the prompt. Make sure your heuristics in 
general is aligned with the given examples. Pay attention to the examples, 
especially because some properties have greater effects when its values are lower.
* You should output CSV files titled "test_pred_A.csv" and "test_pred_B". 
The header is "molecule_a, molecule_b, is_a_greater".
* Only output 0 or 1 for "is_a_greater". 0 means that the property value of 
molecule A is less than that of Molecule B. 1 means that the property value of 
molecule A is greater than that of Molecule B
* In addition to the output CSV file, you should output the first and last 
molecule pairs, in both CSV A and CSV B, to the chat. Briefly explain your 
answer for these three molecule pairs.

# Examples

<user_query>
The property of interest is lipophilicity, the ability of a drug to dissolve in a 
lipid (e.g. fats, oils) environment.
Examples:
CC(C)Cn1c(=O)n(C)c(=O)c2c(C(=O)N3CC[C@@H](O)C3)c(Oc3cccc4ccccc34)sc21,
CC(C)(C(=O)O)c1ccc(C(O)CCCN2CCC(C(O)(c3ccccc3)c3ccccc3)CC2)cc1,
1
O=C(Nc1ccccc1Cl)c1cc[nH]n1,
O=C(Nc1ccc(Nc2ccc(NC(=O)c3ccccc3)c3c2C(=O)c2ccccc2C3=O)c2c1C(=O)c1ccccc1C2=O)c1ccccc1,
0
Cn1c2ccccc2c2cc(NC(=O)CCc3ccncc3)ccc21,
Cc1cc(OCCCS(C)(=O)=O)cc(C)c1-c1cccc(COc2ccc3c(c2)OC[C@H]3CC(=O)O)c1,
1
Cc1ccnc2nc(C(=O)Nc3nccs3)nn12,
Cc1cc(OCCCS(C)(=O)=O)cc(C)c1-c1cccc(COc2ccc3c(c2)OC[C@H]3CC(=O)O)c1,
0
O=C(NCc1ccc(OC(F)(F)F)cc1)C1c2ccccc2C(=O)N1CC1CC(F)(F)C1,
COCCCOc1ccnc(C[S+]([O-])c2nc3ccccc3[nH]2)c1C,
1
</user_query>
\end{verbatim}
\newpage
\subsection{User Study}
\begin{figure}
    \centering
    \includegraphics[width=0.95\linewidth]{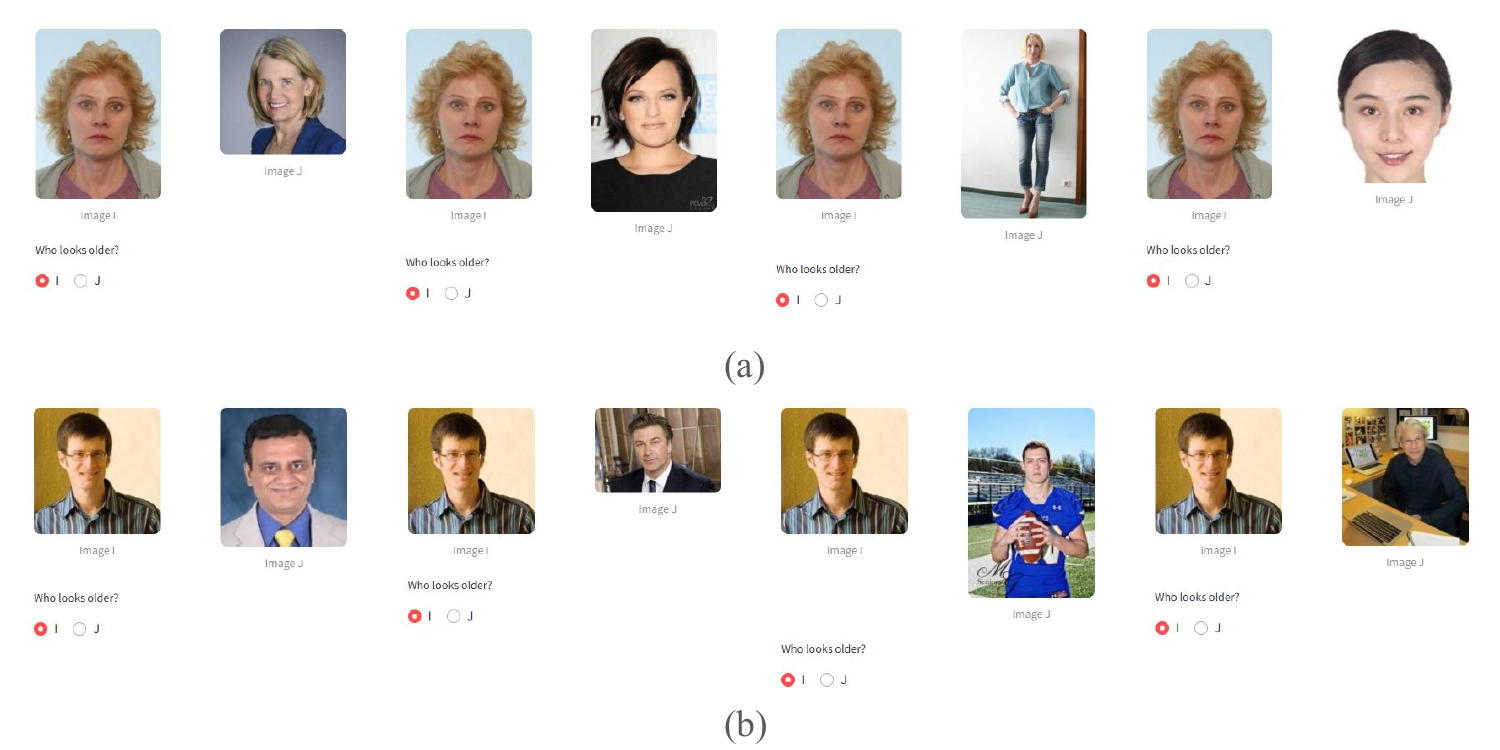}
    \caption{Partial interface used in the user study. Participants first estimate the age of each test individual, then make pairwise judgments on who appears older between the test individual and reference individuals of the same UTKFace gender label.}
    \label{fig:UI_user_study}
\end{figure}

The user study uses age estimation on the UTKFace dataset as a regression task. 
Six participants estimate the ages of 17 individuals from facial images. 
Each test individual is then paired with 15 disjoint reference individuals, and participants judge who appears older in each pair based solely on facial appearance. 
To prevent fatigue, the total number of comparisons is limited to $17 \times 15 = 255$.
A screenshot of the Streamlit app used in the study is shown in Figure~\ref{fig:UI_user_study}.

\section{More Comparisons with the Projection-based Approach}

\begin{figure}[htbp]
    \centering
    \begin{subfigure}[b]{0.32\textwidth}
        \includegraphics[width=\textwidth]{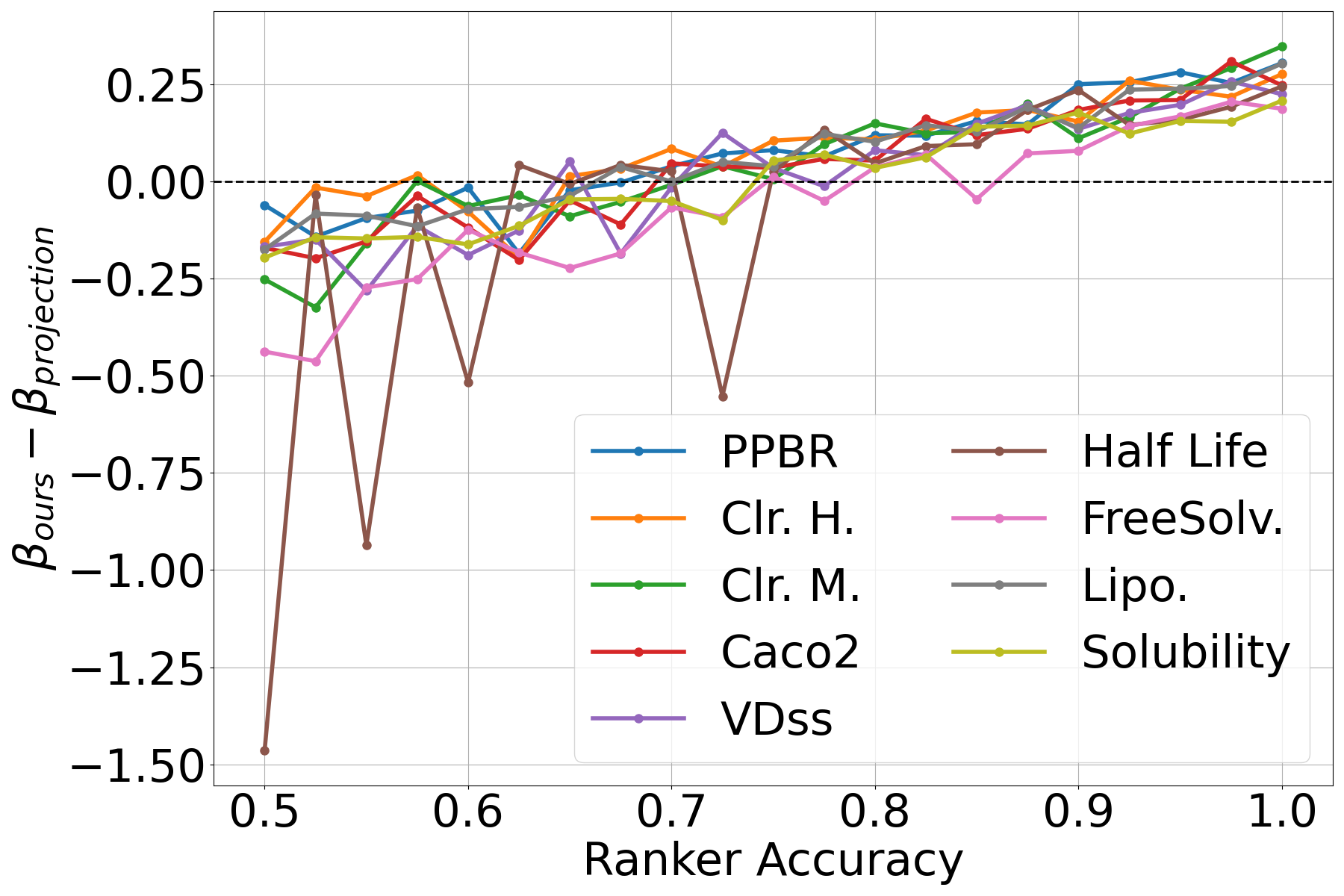}
        \caption{k = 3}
    \end{subfigure}
    \hfill
    \begin{subfigure}[b]{0.32\textwidth}
        \includegraphics[width=\textwidth]{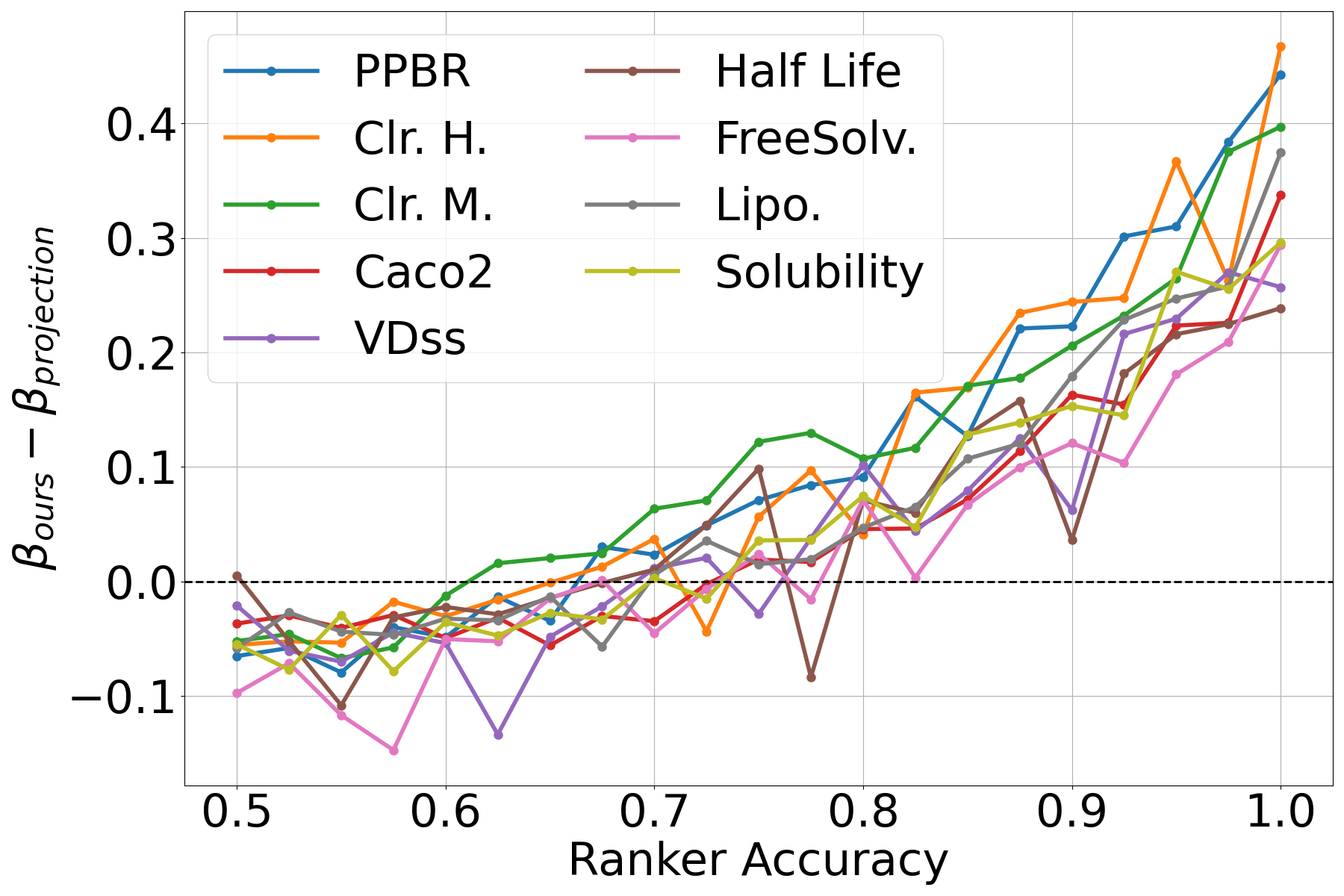}
        \caption{k = 5}
    \end{subfigure}
    \hfill
    \begin{subfigure}[b]{0.32\textwidth}
        \includegraphics[width=\textwidth]{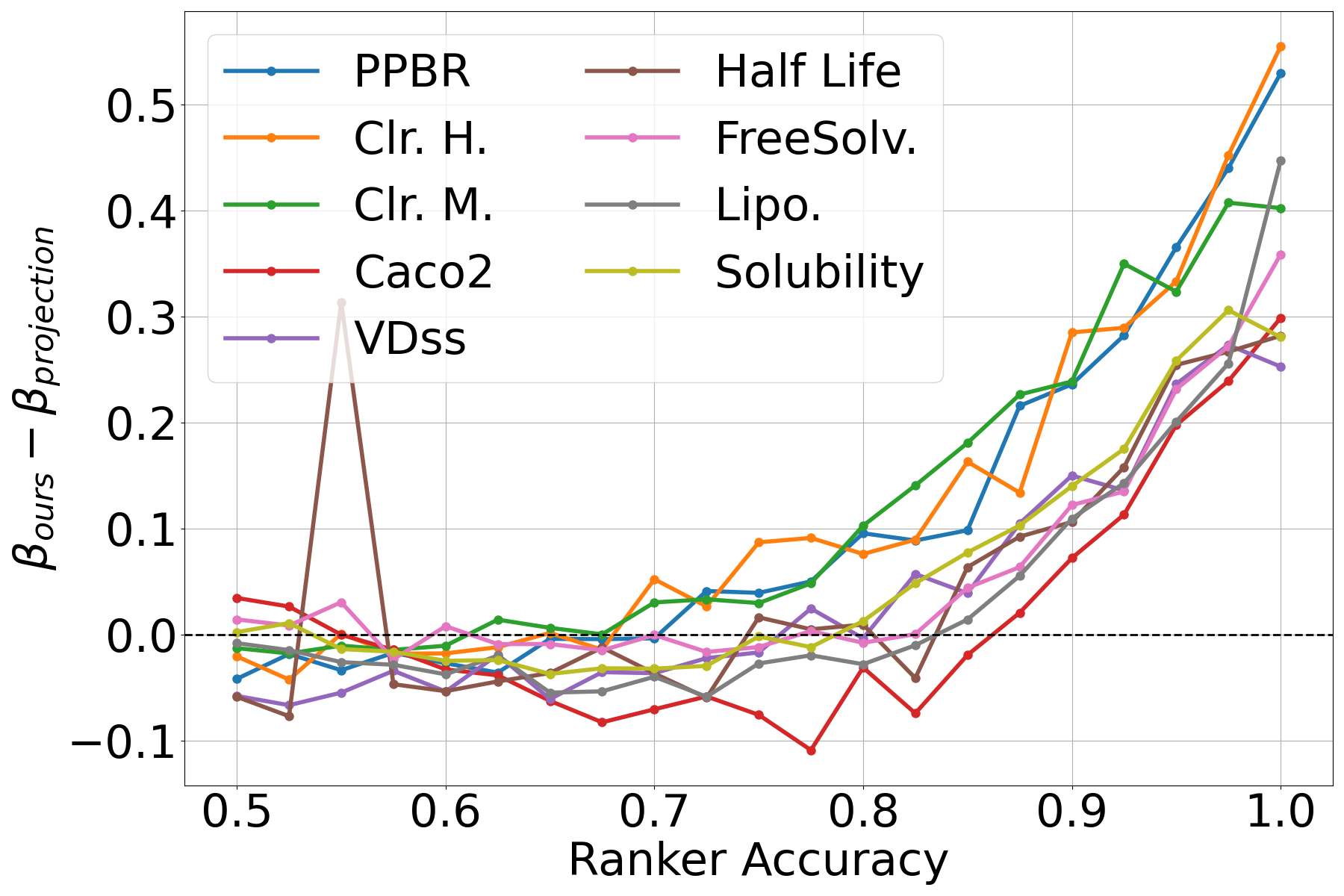}
        \caption{k = 7}
    \end{subfigure}

    \vspace{0.5cm}

    \begin{subfigure}[b]{0.32\textwidth}
        \includegraphics[width=\textwidth]{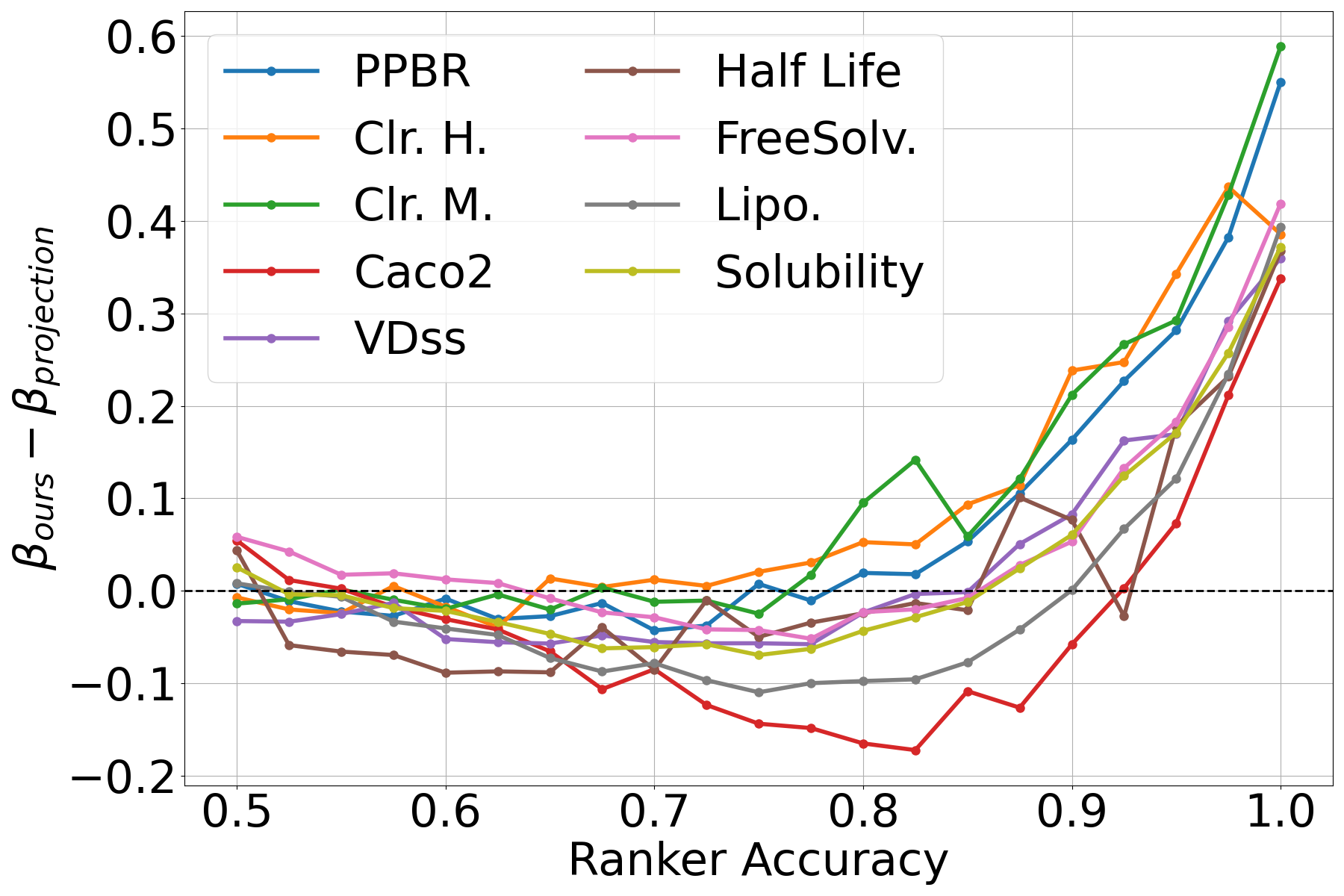}
        \caption{k = 10}
    \end{subfigure}
    \hfill
    \begin{subfigure}[b]{0.32\textwidth}
        \includegraphics[width=\textwidth]{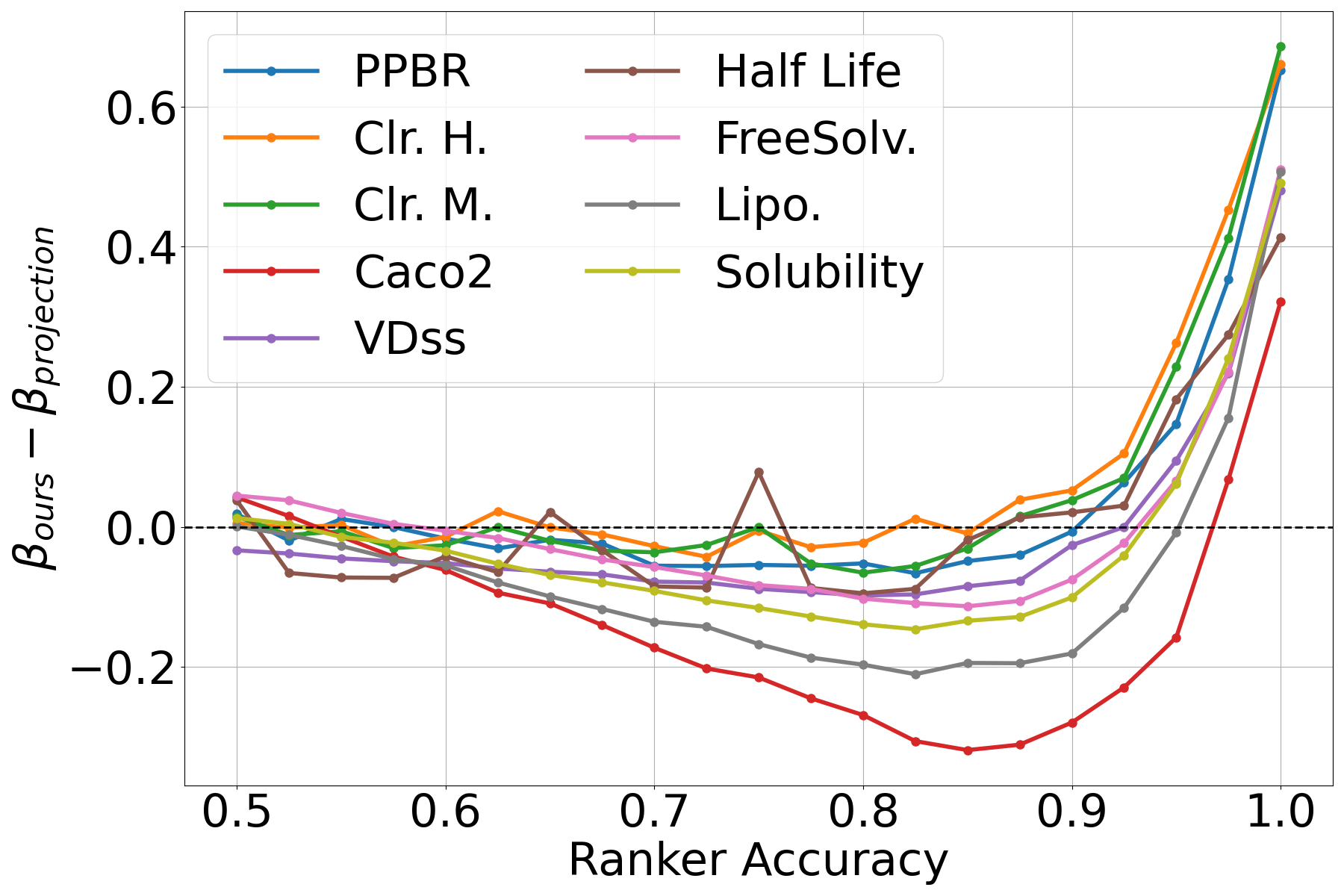}
        \caption{k = 20}
    \end{subfigure}
    \hfill
    \begin{subfigure}[b]{0.32\textwidth}
        \includegraphics[width=\textwidth]{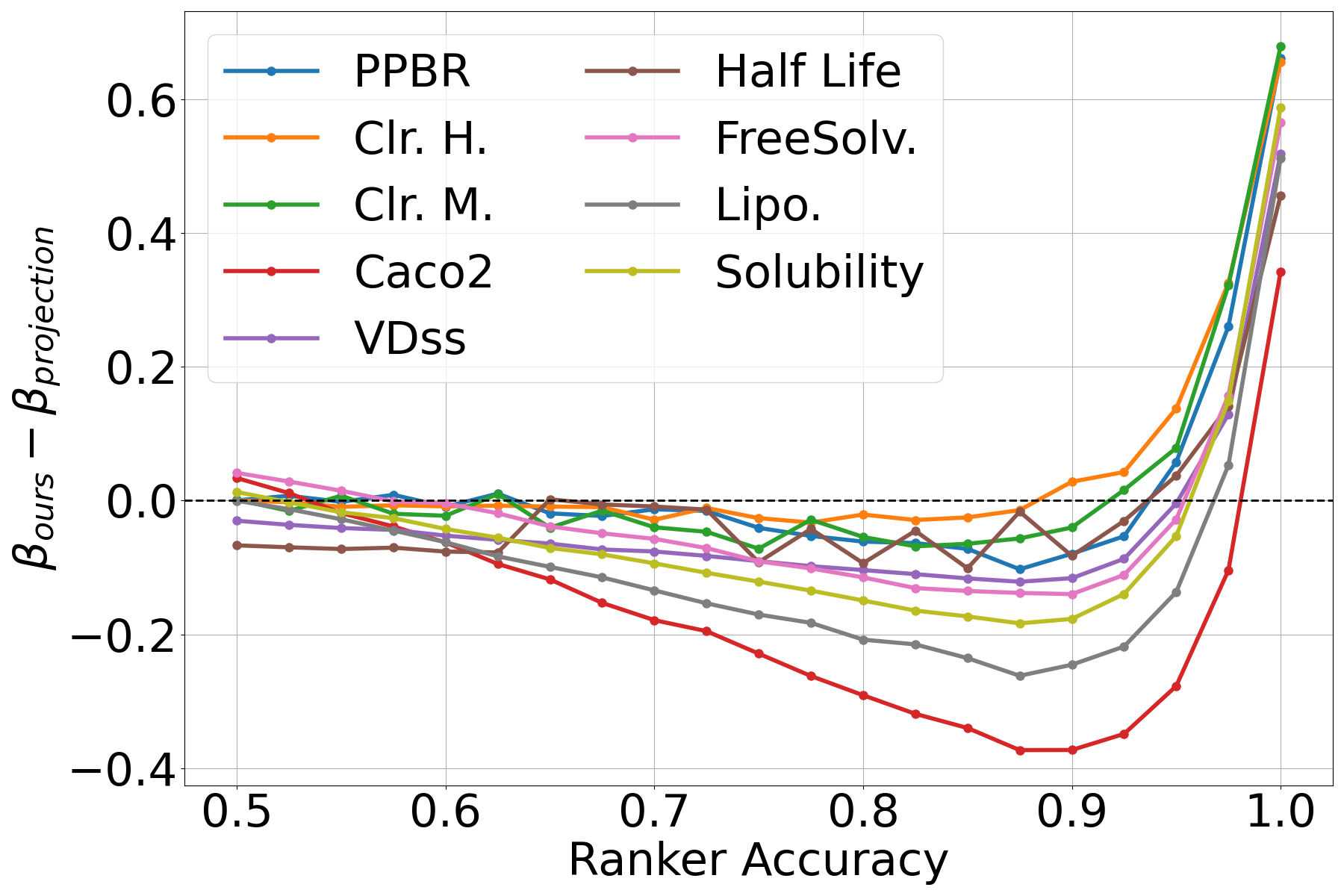}
        \caption{k = 30}
    \end{subfigure}

    \vspace{0.5cm}

    \caption{Comparison of \RankRefine\ and the projection-based approach across varying reference sample sizes $k$. The projection-based method performs better with high ranker accuracy or very few references, while \RankRefine\ excels when accuracy is below 70\% for $k = 7$, 80\% for $k = 10$, 90\% for $k = 20$, and 95\% for $k = 30$. Since \RankRefine\ can leverage general-purpose LLMs, collecting more pairwise rankings can be done efficiently as long as the reference set is sufficiently large.}
    \label{fig:projectionVsOurs}
\end{figure}

Figure \ref{fig:projectionVsOurs} highlights how performance varies between \RankRefine\ and the projection-based approach as the number of reference samples increases. 
While the projection-based method benefits from high-accuracy rankers or minimal references, \RankRefine\ demonstrates stronger robustness in lower-accuracy regimes, particularly as the number of reference samples $k$ increases.

\section{Ablation: Biased Regressor}
We inject constant bias $B$ into an otherwise unbiased regressor, where the magnitude of the offset is measured as a percentage of the ground truth labels standard deviation. 
RankRefine offers improvement even when the regressor is biased (e.g., 60\% of the SD of the ground truth labels)
\begin{table}[ht]
\centering
\caption{RankRefine performance, measured in $\beta$ (lower is better), on a biased regressor. }
\label{tab:ablation_biasRegressor}
\begin{tabular}{lccccc}
    \toprule
    \multicolumn{1}{c}{\textbf{Ranker Accuracy}} & \multicolumn{1}{c}{\textbf{50\%}} & \multicolumn{1}{c}{\textbf{55\%}} & \multicolumn{1}{c}{\textbf{60\%}} & \multicolumn{1}{c}{\textbf{65\%}} & \multicolumn{1}{c}{\textbf{70\%}} \\
    \cmidrule(lr){1-6}
    \textit{\textbf{Bias (B) = 0\% SD}} & 0.8911 & 0.8436 & 0.8694 & 0.8668 & 0.8949 \\
    \textit{\textbf{B = 10\% SD}} & 0.8995 & 0.8800 & 0.8623 & 0.8602 & 0.8969 \\
    \textit{\textbf{B = 20\% SD}} & 0.9084 & 0.8869 & 0.8704 & 0.8704 & 0.9047 \\
    \textit{\textbf{B = 30\% SD}} & 0.9224 & 0.8997 & 0.8844 & 0.8861 & 0.9192 \\
    \textit{\textbf{B = 40\% SD}} & 0.9409 & 0.9172 & 0.9059 & 0.9071 & 0.9395 \\
    \textit{\textbf{B = 50\% SD}} & 0.9645 & 0.9403 & 0.9329 & 0.9343 & 0.9654 \\
    \textit{\textbf{B = 60\% SD}} & 0.9942 & 0.9702 & 0.9649 & 0.9670 & 0.9967 \\
    \textit{\textbf{B = 70\% SD}} & 1.0291 & 1.0062 & 1.0022 & 1.0049 & 1.0326 \\
    \bottomrule
\end{tabular}
\end{table}

\section{Ablation: Biased Sampling}
We restrict the reference set range to increasingly narrow subsets of the ground truth label range ([0, 1]), simulating clustered or biased samples. 
The reference set bias, $RB = x\%$, means that the reference labels cover only $(100-x)\%$ of the full ground-truth range.
For example, an $RB$ = 10\% means the reference labels range is [0.05, 0.95].
Best performance is achieved when the reference set spans the entire range of the ground truth labels, but clustered reference sets still provide substantial regression improvements.
\begin{table}[ht]
\centering
\caption{RankRefine performance, measured in $\beta$, with non-uniform sampling of the reference set.}
\label{tab:ablation_biasSampling}
\begin{tabular}{lccccc}
    \toprule
    \multicolumn{1}{c}{\textbf{Ranker Accuracy}} & \multicolumn{1}{c}{\textbf{50\%}} & \multicolumn{1}{c}{\textbf{55\%}} & \multicolumn{1}{c}{\textbf{60\%}} & \multicolumn{1}{c}{\textbf{65\%}} & \multicolumn{1}{c}{\textbf{70\%}} \\
    \textit{\textbf{Reference Set Bias (RB) = 0\% range}} & 0.8828 & 0.8706 & 0.8343 & 0.8443 & 0.8598 \\
    \cmidrule(lr){1-6}
    \textit{\textbf{RB = 10\% range}} & 0.9034 & 0.8731 & 0.8493 & 0.8529 & 0.8768 \\
    \textit{\textbf{RB = 20\% range}} & 0.9078 & 0.8753 & 0.8366 & 0.8694 & 0.9002 \\
    \textit{\textbf{RB = 30\% range}} & 0.9076 & 0.8701 & 0.8439 & 0.8733 & 0.9280 \\
    \textit{\textbf{RB = 40\% range}} & 0.9143 & 0.8767 & 0.8618 & 0.8824 & 0.9414 \\
    \textit{\textbf{RB = 50\% range}} & 0.9253 & 0.8786 & 0.8640 & 0.8769 & 0.9480 \\
    \textit{\textbf{RB = 60\% range}} & 0.9349 & 0.8858 & 0.8712 & 0.8929 & 0.9546 \\
    \textit{\textbf{RB = 70\% range}} & 0.9385 & 0.8919 & 0.8770 & 0.8957 & 0.9597 \\
    \textit{\textbf{RB = 80\% range}} & 0.9551 & 0.8984 & 0.8798 & 0.8943 & 0.9558 \\
    \textit{\textbf{RB = 90\% range}} & 0.9664 & 0.9042 & 0.8842 & 0.9035 & 0.9631 \\
    \bottomrule
    \end{tabular}
\end{table}

To test the impact of distribution shift between reference and query, we created two challenging scenarios where the reference labels and query labels are drawn uniformly from different range values:
\begin{itemize}
    \item \textit{Disjoint} distributions: reference labels in [-1, 0] or [1, 2], query labels in [0, 1]. We averaged the results of the two cases, that is, between {ref = [-1, 0], query = [0, 1]} and {ref = [1, 2], query = [0, 1]}.
    \item \textit{Partial overlap}: reference labels in [-0.5, 0.5] or [0.5, 1.5], query labels in [0, 1]. We averaged the results of the two cases, that is, between {ref = [-0.5, 0.5], query = [0, 1]} and {ref = [0.5, 1.5], query = [0, 1]}.
\end{itemize}  
The results in \ref{tab:shifted_sampling} shows that RankRefine still improves the base regressor when the ranker is moderately accurate, even when the distribution of the reference labels is different from the query.
\begin{table}[ht]
\centering
\caption{RankRefine performance, measured in $\beta$,  for disjoint and partially overlapping reference set.}
\label{tab:shifted_sampling}
    \begin{tabular}{lccccc}
    \toprule
    \multicolumn{1}{c}{\textbf{Ranker Accuracy}} & \multicolumn{1}{c}{\textbf{50\%}} & \multicolumn{1}{c}{\textbf{55\%}} & \multicolumn{1}{c}{\textbf{60\%}} & \multicolumn{1}{c}{\textbf{65\%}} & \multicolumn{1}{c}{\textbf{70\%}} \\
    \cmidrule(lr){1-6}
    \textit{\textbf{Disjoint}} & 1.2906 & 1.1629 & 1.0170 & 0.9437 & 0.9219 \\
    \textit{\textbf{Partial Overlap}} & 1.0160 & 0.9465 & 0.8700 & 0.8607 & 0.8898 \\
    \bottomrule
    \end{tabular}
\end{table}
\end{document}